\pdfoutput=1
\documentclass[11pt]{article}

\usepackage[]{acl}

\usepackage{times}
\usepackage{latexsym}

\usepackage[T1]{fontenc}
\usepackage[utf8]{inputenc}

\usepackage{microtype}
\usepackage{booktabs}
\usepackage{tabularx}
\usepackage{amsmath}
\usepackage{multirow}
\usepackage{graphics}
\usepackage{graphicx}
\usepackage{amsfonts}
\usepackage{CJK}
\usepackage{subfigure}
\usepackage{soul}
\usepackage{color}
\usepackage{xcolor}
\newcommand{\hlred}{\colorlet{c}{red!20}\sethlcolor{c}\hl}

\title{A Corpus for Understanding and Generating Moral Stories}

\author{Jian Guan, Ziqi Liu, Minlie Huang\Thanks{~Corresponding author} \\
The CoAI group, DCST;
Institute for Artificial Intelligence; State Key Lab of \\Intelligent Technology and Systems; Beijing National Research Center for \\Information Science and Technology; Tsinghua University, Beijing 100084, China.\\
  \texttt{\{j-guan19, liuzq19\}@mails.tsinghua.edu.cn},\\ \texttt{aihuang@tsinghua.edu.cn}}

\begin{document}
\maketitle

\begin{abstract}
Teaching morals is one of the most important purposes of storytelling. An essential ability for understanding and writing moral stories is bridging story plots and implied morals. Its challenges mainly lie in: (1) grasping knowledge about abstract concepts in morals, (2) capturing inter-event discourse relations in stories, and (3) aligning value preferences of stories and morals concerning good or bad behavior. In this paper, we propose two understanding tasks and two generation tasks to assess these abilities of machines. We present \textsc{storal}, a new dataset of Chinese and English human-written moral stories. We show the difficulty of the proposed tasks by testing various models with automatic and manual evaluation on \textsc{storal}. Furthermore, we present a retrieval-augmented algorithm that effectively exploits related concepts or events in training sets as additional guidance to improve performance on these tasks.
%Finally, we propose decoding strategies that effectively combine multiple expert models to significantly improve the quality of generated actions, consequences, and norms compared to strong baselines, e.g. though abductive reasoning.
\end{abstract}

\section{Introduction}

%Storytelling is an effective approach for education~\cite{}. 

%Morals have long been one of the primary purpose of storytelling, especially in children's literature such as fables and fairy tales, which often imply abstract morals through concrete events for children's education.

Stories play an essential role in one's moral development~\cite{vitz1990use}. For example, individuals usually learn morals from life experiences or literature such as fables %and fairy tales, 
and tell their morals by representing their lived experience in a narrative  form~\cite{tappan1989stories}. Accordingly, it is a crucial ability for humans to bridge abstract morals and concrete events in stories. However, this ability has not yet been investigated for machines. 

\iffalse
\begin{figure}[!ht]
\includegraphics[width=\linewidth]{graph/intro_example.pdf}% 1\linewidth
  \caption{An example in \textsc{storal}.}%, the discourse structure of the story~(Bottom Left) and the association between the story and the moral~(Bottom Right).}
  \label{tab:example_story}
\end{figure}
\fi

\begin{table}[!t]
\small
    \centering
    \begin{tabular}{p{207pt}}
    \toprule
    \textbf{Stories:} 
    Four cows lived in a forest near a meadow. They were good friends and did everything together. They grazed together and stayed together, because of which no tigers or lions were able to kill them for food. \\~~~~But one day, the friends fought and each cow went to graze in a different direction. A tiger and a lion saw this and decided that it was the perfect opportunity to kill the cows. They hid in the bushes and surprised the cows and killed them all, one by one.\\
    \midrule
    \textbf{Morals:} Unity is strength.\\
    \bottomrule 
    \end{tabular}
    \caption{An example in \textsc{storal}}
    \label{tab:example_story}
\end{table}

    %Four cows lived in a forest near a meadow. They were good friends and did everything together. They grazed together and stayed together, because of which no tigers or lions were able to kill them for food. \\~~~~But one day, the friends fought and each cow went to graze in a different direction. A tiger and a lion saw this and decided that it was the perfect opportunity to kill the cows. They hid in the bushes and surprised the cows and killed them all, one by one.\\
    
    %\textbf{Moral:} Unity is strength.\\

There have been many tasks proposed for evaluating story understanding and generation, including story ending selection~\cite{mostafazadeh2016corpus} %story completion~\cite{DBLP:conf/ijcai/Wang019b} 
and story generation from short prompts~\cite{fan2018hierarchical}. Unlike these tasks, which focus on reasoning plots from context, we emphasize the ability to associate plots with implied morals. As exemplified in Table~\ref{tab:example_story}, the challenges mainly lie in (1) grasping knowledge about abstract concepts~(e.g., {``unity,''} {``strength''}) and relations among them~(e.g., {``is''}) in morals; %, which may become more challenging for morals with rhetoric~(e.g., {``failure breeds success''}); 
(2) capturing inter-event discourse relations in stories (e.g., the contrast between endings of the {``cows''} when they are ``united'' and ``divided''); and (3) aligning value preferences~\cite{jiang2021delphi} of stories and morals %which requires a strong ability of reasoning the authors' intention 
(e.g., the story implies support for {``unity''}, not opposition,  which agrees with {``is strength''} in the moral). To test these abilities of machines, we propose two understanding tasks and two generation tasks. Both understanding tasks require selecting the correct moral from several candidates given a story. And they have respective candidate sets for testing machines in two aspects, including concept understanding~(\textsc{moCpt} for short) and preference alignment~(\textsc{moPref} for short). The generation tasks require concluding the moral of a story~(\textsc{st2mo} for short), and conversely generating a coherent story to convey a moral~(\textsc{mo2st} for short).
%, %and evaluate the ability of existing models, 
%we formulate two understanding tasks which require selecting the correct moral from several candidates given a story, and two generation tasks including concluding a moral for a story, and conversely writing a coherent story to convey a moral. Furthermore, we introduce a new dataset named \textsc{storal} composed of Chinese and English moral stories to support the evaluation.

Furthermore, we collected a new dataset named \textsc{storal} composed of 4k Chinese and 2k English human-written stories paired with morals through human annotation to  %as a resource 
address the above challenges. We call the Chinese dataset \textsc{storal-zh} and the English dataset \textsc{storal-en}, respectively. And we construct datasets for the proposed tasks based on \textsc{storal}. Our focus of morals is on the %lessons in stories such as 
social set of standards for good or bad behavior and character, or the quality of being right, honest or acceptable~\cite{ianinska2006morals}.  
We conduct extensive experiments on the proposed tasks. Furthermore, we present a retrieval-augmented algorithm to improve model performance by retrieving related concepts or events from training sets as additional guidance. %We summarize  our contribution as follows:
%\noindent\textbf{I.} We propose two understanding tasks and two generation tasks for evaluating the ability to associate story plots and implied morals. And we present \textsc{storal}, a new dataset of Chinese and English moral stories, for testing theses tasks.
%\noindent\textbf{II.} We propose 
However, the experiment results demonstrate that existing models still fall short of understanding and generating moral stories, which requires a better modeling of discourse and commonsense relations among concrete events and abstract concepts
%better semantic representations of events and abstract concepts, and deeper modeling of commonsense and discourse relations among them
\footnote{All data and evaluation scripts are available at \url{https://github.com/thu-coai/MoralStory}.}.
%richer semantic representation of events together with deeper levels of modeling the semantic space of stories.
%We believe that \textsc{storal} and  switching to the two proposed tasks as the empirical evaluation framework for story understanding and generation can help direct the field to a new direction of NLP.

%unveiling the moral of a story  requires a wealth of association between 

%Morals were one of the main purposes of literature during 1780–1830, especially in children's literature. Part of the reason for this was the writings of John Locke and Jean-Jacques Rousseau in the 18th century, which brought attention to children as an audience for literature. Following in their line of thought, Thomas Day (1748–1789) wrote Sandford and Merton, elevating the outstanding morals of one young boy above the rapscallion nature of another. Maria Edgeworth (1776–1849) was another prominent author of moral tales, writing about how a wise adult can educate a child; one of her more famous stories is "The Purple Jar". During this time, the theme of "a young heroine or hero gaining wisdom and maturity was taken up by many other writers".[2]

%The ability of children to derive moral lessons from stories and visual media develops around the age of 9 or 10 years.[3]

%为什么要做哲理故事生成的数据集？
%哲理故事相比于之前的故事有哪些新的挑战？
%我们的数据集怎么能够支持模型应对这些挑战？
%总结我们的贡献：
%1. First study to associate the narrative plots with abstract morals
%2. A Chinese dataset and an English dataset
%3. Extensive experiments to benchmark existing models

\section{Related Work}
\paragraph{Story Datasets} %Prior studies on story understanding and generation have mainly focused on 
ROCStories~\cite{mostafazadeh2016corpus} and WritingPrompts~\cite{fan2018hierarchical} are two frequently used story datasets in related studies. The former consists of artificial five-sentence stories regarding everyday events, while the latter contains fictional stories of 1k words paired with short prompts. Besides, some recent works collected extra-long stories %from public resources as benchmark datasets 
such as {roleplayerguild}~\cite{louis2018deep}, PG-19~\cite{Rae2020Compressive},  and \textsc{storium}~\cite{akoury2020storium}. \citet{guan2021lot} proposed a collection of Chinese stories. These stories usually aim to narrate a coherent event sequence but not convince readers of any morals. %In contrast, \textsc{storal} focuses on those stories with annotation of morals that they are trying to convey. %Such stories usually have a specific discourse structures from previous narrative ones. For example, the stories may use a structured set of premises to support their argument and demolish potential counterarguments~\cite{jurafsky2000speech}. %the claim together with a structured set of premises that support the argument and demolish potential counterarguments.

\paragraph{Story Understanding and Generation}
There have been many tasks proposed for evaluating story understanding and generation. %from following perspectives. 
Firstly, some works tested the machinery commonsense reasoning ability regarding inter-event causal and temporal relations through story ending selection~\cite{mostafazadeh2016corpus}, story ending generation~\cite{guan2019story} and story completion~\cite{DBLP:conf/ijcai/Wang019b}. %focused on {commonsense reasoning} in terms of causal and temporal relations between events. 
Secondly, a series of studies focused on the coherence of story generation~\cite{fan2018hierarchical, yao2019plan, guan2020knowledge}. %Another line is learning to judge the coherence of a story  such as evaluating story generation~\cite{guan2020union} and predicting the correct position of a sentence in a story~\cite{chen2019evaluation}. 
Another line of works concentrated on {controllability} to impose specified attributes into story generation. These attributes involved %keywords~\cite{DBLP:conf/emnlp/XuPSPFAC20}, 
outlines~\cite{rashkin2020plotmachines}, emotional trajectories~\cite{brahman2020modeling} and story styles~\cite{kong-etal-2021-stylized}. Our tasks %serve as a comprehensive benchmark 
investigate not only the above aspects but also the ability to understand abstract concepts and reason value preferences of stories. %and use persuasive writing techniques.

%condense
A task similar to \textsc{st2mo} is text summarization~\cite{finlayson2012learning} since both tasks require generating a short text to condense crucial information of a long text. But summarization requires reorganizing a few words of the original text instead of concluding a character-independent moral. For example, a plausible summary of the story in Table~\ref{tab:example_story} is {``Four cows were killed by two tigers and a lion''} (generated by BART$_{\rm Large}$~\cite{bart} fine-tuned on a summarization dataset XSUM~\cite{narayan2018don}), which includes specific characters and events of the original story. Moreover, \textsc{mo2st} is similar to persuasive essay generation~\cite{stab2017parsing}, which also requires conveying a viewpoint in generated texts. However, %poetry generation requires exhibiting explicit structured rules, and 
persuasive essays usually convince readers by directly presenting arguments but not narrating a story. 

\paragraph{Morals} \citet{haidt2004intuitive} provided a theoretical framework named Moral Foundations Theory~(MFT) to summarize five basic moral foundations such as ``Care/Harm,'' ``Fairness/Cheating,'' etc. Based on the theory, recent studies have explored to classify the moral foundations of partisan news~\cite{fulgoni2016empirical}, tweets~\cite{johnson2018classification, doi:10.1177/1948550619876629}, and crowd-sourced texts~\cite{pavan2020morality}. And \citet{volkova2017separating} proposed identifying suspicious news based on the features of moral foundations. However, we focus on morals which are free-form texts far beyond the scope of the five categories in MFT. In addition,  recent studies proposed multiple datasets for machine ethics research such as SBIC~\cite{sap-etal-2020-social}, Social Chemistry~\cite{forbes-etal-2020-social}, Moral Stories~\cite{emelin2020moral}, ETHICS~\cite{hendrycks2021aligning} and Scruples~\cite{lourie2021scruples}. But these datasets focus more on %machine ethics and AI safety in terms of 
how machines behave ethically in some scenario, while \textsc{storal} emphasizes the ability to conclude the moral implied by a story. Moreover, most cases in these datasets consist of short texts of descriptive ethical behavior, typically in the form of one sentence. %\citet{emelin2020moral}  and \citet{jiang2021delphi} explored to distinguish moral actions from immoral ones in a form of one sentence, 
In contrast, \textsc{storal} provided longer and more context-specific stories for moral understanding.

\begin{figure*}[!t]
\centering
\includegraphics[width=0.9\linewidth]{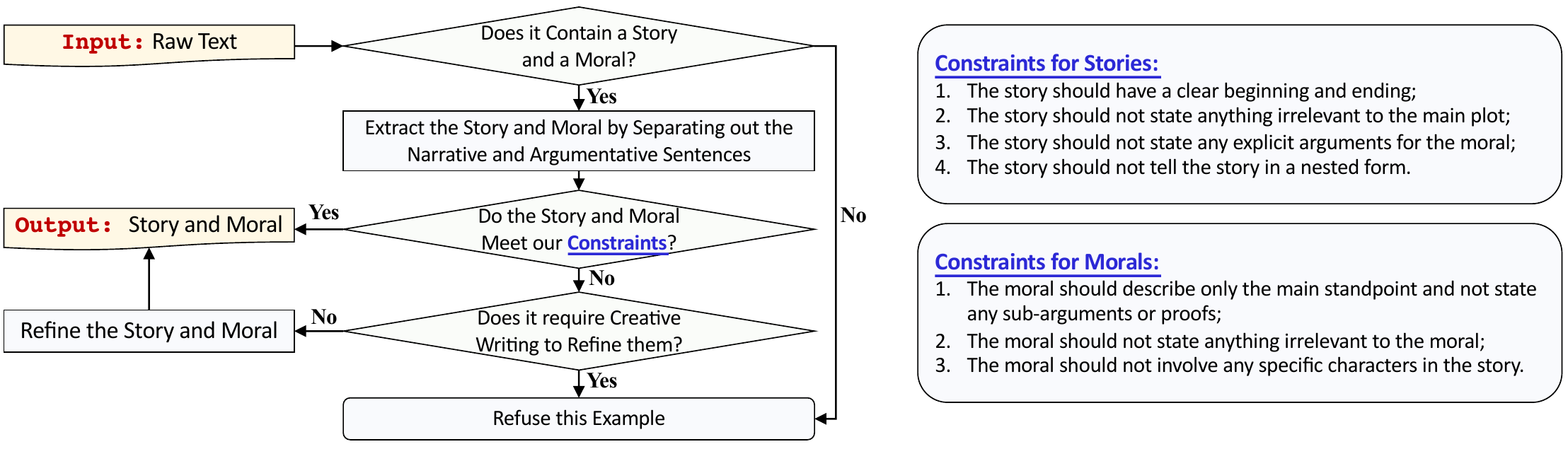}  
\caption{The pipeline of human annotation for constructing \textsc{storal} (Left) and our constraints~(Right).}
  \label{tab:pipeline}
\end{figure*}

\section{\textsc{storal} Dataset}
We collected \textsc{storal} from multiple web pages of moral stories. %paired with morals professionally written for them, typically by their authors. 
All stories are allowed to use and redistribute for research and have been reviewed by the website editors as stated on the pages. We show the full list of links to these pages in Section~\ref{source}. After de-duplication, we collected 19,197 Chinese and 2,598 English raw texts. Then we adopted human annotation for decoupling the story and moral in each raw text. Due to resource limitations, we only constructed 4,209 Chinese and 1,779 English story-moral pairs.
%Finally we selected 6/14 websites as the data source for STORAL-ZH/EN, respectively. We will add a total list of the websites in our revision.
We will first show the details of human annotation, then present the topic analysis and statistics of \textsc{storal}, and finally describe the details of dataset construction for the proposed tasks.  %which is professionally written, typically by the author of the article.
% 网站列表appendix

%We've collected several stories with moral in both Chinese and English as training data and valid data. We divide the stories into two parts: story and moral sentence. Moral sentence contains a common moral or ethic which could be reflected in story. Besides, moral sentence should not be heavily dependent on story plot so it couldn't contain the name that appears in the matched story. 
\subsection{Human Annotation}

To narrow down our focus, we %follow \citet{mostafazadeh2016corpus} to 
define a story as a series of coherent events involving several inter-related characters, and implies support or opposition of some behavior. Such a definition constrains the story to %have a clear beginning and ending with something that happens in between and aim to 
exhibit a moral without any explicit arguments. %\footnote{We believe that all the stories that we collect aims to exhibit certain morals  since they are paired with corresponding moral sentences. }.
And we define a moral as a judgment to describe what the story implies concerning good or bad behavior. Note that we do not require morals in \textsc{storal} to be always reflective of normatively virtuous behavior. We emphasize that the morals should align with the story. Then, a key issue is how to extract the story and moral from a raw text. We observe that there are no markers such as ``The story tells us'' to separate the story and moral in most cases. The moral may be tightly weaved into the plot (e.g., included in a dialogue). Therefore, we adopted human annotation for this extraction task. We hired a commercial team to annotate \textsc{storal-zh}. All annotators are native Chinese speakers and well trained for our task. 
For \textsc{storal-en}, we hired three graduates with good English language proficiency. We did not use AMT since it is inconvenient to train online annotators. Figure~\ref{tab:pipeline} shows the annotation pipeline.

We first ask annotators to judge whether the raw text contains a story and moral and whether they meet our constraints shown in Figure~\ref{tab:pipeline}. We show the examples given to the annotators to inform them of our requirements for stories and morals in Section~\ref{constraints}. If the constraints are not met, we then ask annotators to refine the story and moral.  %to complete the story plots 
%since such a completion task is much more difficult even for humans~(e.g., completing the first example of Table~\ref{tab:st}) and it would also bring challenges to quality control. 
In the refinement stage, %we also instruct annotators to delete noisy words such as links and codes. In addition, 
annotators have to clean up the data with following heuristics: (1) refusing examples which may violate general ethical principles~(e.g., discrimination); (2) deleting noisy words~(e.g., links, codes); %(3) ensuring that there is little English/Chinese words in \textsc{storal-zh}/\textsc{en}; (3) and setting all the punctuation marks as full/half-width for \textsc{storal-zh}/\textsc{en} and deleting . and 
(3) refining the stories and morals to be coherent and formal. %~(e.g., deleting %unnecessary interjections such as {``ah''}, and 
%irregular punctuation marks such as {``$\sim$''}). 
And to ensure the quality of collected data, annotators may refuse to refine the example if it requires much creative writing.
Finally, we review the annotation results and provide detailed feedback to the annotators before approving their submissions. We show an annotation example in Table~\ref{tab:ant_example}.

\begin{table}[!th]
\scriptsize
    \centering
    \begin{tabular}{p{207pt}}
    \toprule
    \textbf{Raw Text:} A man who\textit{{wWw.xxx.c0M}}lived a long time ago believed that he could read the future in the stars. He called himself an Astrologer, and spent his time at night gazing at the sky. One evening he was walking along the open road outside the village. His eyes were fixed on the stars. He thought he saw there that the end of the world was at hand, when all at once, down he went into a hole full of mud and water. There he stood up to his ears, in the muddy water, and madly clawing at the slippery sides of the hole in his effort to climb out. 
    His cries for help soon brought the villagers running. As they pulled him out of the mud, one of them said:``You pretend to read the future in the stars, and yet you fail to see what is at your feet! \textit{This may teach you to \textbf{pay more attention to what is right in front of you, and let the future take care of itself.}''``what use is it? '' said another, ``
    to read the stars, when you can't see what's right here on the earth?''}\\
    \midrule
    \textbf{Story:} A man who lived a long time ago believed that he could read $\cdots$ As they pulled him out of the mud, one of them said: ``You pretend to read the future in the stars, and yet you fail to see what is at your feet!''\\
    \midrule
    \textbf{Moral:} Pay more attention to what is right in front of you, and let the future take care of itself.\\
    \bottomrule
    \end{tabular}
    \caption{An example for extracting the story and moral from a raw text. We highlight the words which should be revised in the raw text in \textit{italic}.  And the moral in the raw text is \textbf{bold}. To save space, we replace some events with ``$\cdots$'' in the story.}
    \label{tab:ant_example}
\end{table}

%we construct \textsc{storal-en} by automatically extracting the stories and morals. , which makes it difficult to tell them apart automatically. Therefore, in order to ensure the credibility of \textsc{storal-zh} for model evaluation, we construct the validation and test sets by manually extracting stories and morals from the raw texts. And we carefully devise an automatic approach to construct the training set of \textsc{storal-zh} considering the high cost of manual annotation. We also demonstrate that using the automatic approach can derive similar results with using manual annotation in spite of a little noise. 

\begin{table*}[!t]
\scriptsize
    \centering
    \begin{tabular}{p{180pt}p{250pt}}
    \toprule
    \textbf{Topic Words} & \textbf{Examples}\\
    \midrule
    \begin{CJK}{UTF8}{gbsn}\scriptsize\underline{懂得~(understand)}, 也是~(also), \underline{了解~(know)}, 方法~(method), 收获~(gain), 保护~(protect), 大脑~(brain), \underline{才能~(able)}, 付出~(pay), 进步~(progress)\end{CJK}&\begin{CJK}{UTF8}{gbsn}\scriptsize在犯错的时候我们要\underline{懂得}看全局，要\underline{了解}全局\underline{才能}对事情有定义。(When making mistakes, we must \underline{understand} the overall situation. And we are \underline{able} to have a definition of things only when \underline{knowing} the overall situation.)\end{CJK}\\
    \midrule
    %聪明 都是 学会 应对 情况 胜利 足够 知识 就能 更多
    \begin{CJK}{UTF8}{gbsn}\scriptsize不要~(not), 一定要~(must), 危险~(danger), 时候~(when), \underline{对待~(treat)}, \underline{安全~(safety)}, 千万~(any way), 好好~(well), 学会~(learn), 遇到~(encounter)\end{CJK}&\begin{CJK}{UTF8}{gbsn}\scriptsize生活中也要牢记“\underline{安全}”这两字，在“\underline{安全}”两字面前切不可存有侥幸心理，把\underline{安全}当成儿戏。 (Keep in mind the word ``\underline{safety}'' in your life, and do not take any chances to \underline{treat} \underline{safety} as a joke.) \end{CJK}\\
    \midrule
    %好孩子 告诉 生活 爸爸妈妈 认真 玩耍 睡觉 成为 食物 平常
    \begin{CJK}{UTF8}{gbsn}\scriptsize\underline{事情~(thing)}, 才能~(able), \underline{做好~(do well)}, 优秀~(excellent), \underline{应该~(should)}, 做到~(achieve), 自信~(self-confident), 有所~(somewhat), 无法~(unable), 可能~(may)\end{CJK}&\begin{CJK}{UTF8}{gbsn}\scriptsize\underline{做好}自己\underline{该}做的\underline{事情}，做自己的主人。(\underline{Do} what you \underline{should} do and be your own master.)\end{CJK}\\
    %态度 执着 亲情 成全 本性 亲人 坚持不懈 保证 精力 站起来\
    \midrule
    \begin{CJK}{UTF8}{gbsn}\scriptsize时候~(when), 其实~(actually), 很多~(many), \underline{发现~(discover)}, 希望~(wish), 发生~(happen), 生活~(life), 已经~(already), 伤害~(hurt), 可能~(may)\end{CJK}&\begin{CJK}{UTF8}{gbsn}\scriptsize人要善于自己\underline{发现}自己，而不是老等着别人来发现我们。(We should be good at \underline{discovering} ourselves instead of waiting for others to do.)\end{CJK}\\
    \midrule
    %就是 不是 就会 勇气 感受 觉得 只是 只要你 镜子 接受
    \begin{CJK}{UTF8}{gbsn}\scriptsize\underline{遇到~(encounter)}, 问题~(question), \underline{困难~(difficulty)}, 解决~(solve) 思考~(think), \underline{帮助~(help)}, 时候~(when), 应该~(should), \underline{给予~(give)}, 头脑~(brain)\end{CJK}&\begin{CJK}{UTF8}{gbsn}\scriptsize乐于\underline{助}人，是一种朴实的中国传统美德。每个人都有\underline{遇到困难}的时候，最需要的是别人\underline{给予}的\underline{帮助}。(Being \underline{helpful} is a Chinese traditional virtue. When someone \underline{encounters difficulties}, what he needs most is \underline{help} from others.)\end{CJK}\\
    \midrule
    \midrule
    {good}, \underline{always}, {come}, \underline{believe}, first, \underline{honesty, speak}, world, \underline{around}, act&\textbf{1.} Always be \underline{honest}. \underline{Honesty} is \underline{always} rewarded. \newline\textbf{2.} A liar will not be \underline{believed}, even when he \underline{speaks} the truth. \\%What goes \underline{around comes} around. You do \underline{good}, you will get \underline{good} in return. \underline{Always} be helpful.\\
    \midrule
    % get policy fool go ignore even return feel helpful temptation
    \underline{help}, also, good, need, hope, lose, \underline{carry}, feel, \underline{say}, \underline{self} &\textbf{1.} One should not be \underline{carried} away by what others \underline{say}. Don't be fooled by those who wants to take advantage of you. \newline\textbf{2.} Self \underline{help} is the best \underline{help}. Heaven \underline{helps} those who \underline{help} \underline{themselves}. \\
    \midrule
    %succeed away put remember thank child achieve story mind level
    \underline{friend}, act, \underline{wisely}, moment, \underline{think}, place, \underline{time}, \underline{choose}, great, ability&\textbf{1. }Little \underline{friends} may prove great \underline{friends}.\newline \textbf{2.} One should not panic in difficult \underline{times} and \underline{think wisely}. \\
    \midrule
    %anger pain eye play consequence prove neighbor hear underestimate weak
    \underline{love, care,  parent, respect, always}, value, take, mean, give, one&\textbf{1. }You reap what you sow. Regardless of your relationship with your \underline{parents}, you'll miss them when they're gone from your life. \underline{Always respect, care} for and \underline{love} them. \newline\textbf{2.} Be content with your lot; \underline{one} cannot be first in everything. \\
    \midrule
    %child people remember feel even never sacrifice feeling easy true
    look, {see}, bad, make, turn, \underline{strong}, strength, choice, give, deserve&
    \textbf{1.} The \underline{strong} and the weak cannot keep company. \newline\textbf{2.} It is easy to despise what you cannot get.\\%Distrust interested advice.  \\
    %Often we misunderstand the signs. God has given everyone a strength and ability to do their part. Always learn to see the things in a positive way and see yourself in a strong position to help the ones who need. Don't go for the easy choice. Make a right choice.\\
    %part right need company positive leap position aspect keep learn
    \bottomrule
    %生活 不是 经常 后悔 难免 毫无 想办法 沟通 需要 完全
\\
    \end{tabular}
    \caption{Topic words and examples for \textsc{storal-zh}~(top) and \textsc{storal-en}~(bottom). We underline the topic words that occurs in the examples.} %We show only one example for each topic in \textsc{storl-zh} due to space limitation.}
    \label{tab:lda_example}
\end{table*}

\subsection{Topic Analysis}\label{topic_ana}
To provide insight into the taxonomy of morals within \textsc{storal}, we adopt LDA~\cite{blei2003latent} for topic modeling of morals. Let $B$ denote the number of topics and $V$ denote the vocabulary size. Based on the variational parameter for topic word distribution $\beta\in\mathbb{R}^{B\times V}$, we determine $B$ as the minimum value that makes the following formula holds true for any $b\in \{1, 2, \cdots, B\}$: 
\begin{align*}
s_b=\frac{\sum_{v\in{\mathcal{V}_b^{(k)}}}\beta_{bv}}{\sum_{v=1}^V\beta_{bv}}\ge h,\\
\mathcal{V}^{(k)}_b = \text{argmax}_{\mathcal{V}^{*(k)}_b}\sum_{v\in\mathcal{V}^{*(k)}_b}\beta_{bv},
\end{align*}
where $\beta_{bv}$ is the element at the $b$-th row and $v$-th column of $\beta$, $k\in\{1,2,\cdots,V\}$ is the size of the top-$k$ vocabulary $\mathcal{V}^{(k)}_b$, and $h\in[0,1]$ is a predefined threshold. $s_b$ is used to measure the specificity of the $b$-th topic. Intuitively, the larger $s_b$, the more specific the topic. We set $k$ to 20 and $h$ to 0.5.
%the sum of top 20 values in each row of $\beta$ accounts for more than 50\% of the total. 
Finally, we derive 40/24 topics for \textsc{storal-zh}/\textsc{storal-en}, respectively. And the minimum proportion of examples of one topic is 1.6\%/3.2\% for \textsc{storal-zh}/\textsc{storal-en}, respectively. 

Table~\ref{tab:lda_example} shows the topic words in $\mathcal{V}^{(10)}$ of each topic and two morals assigned to each topic with the highest probabilities for the five topics with the largest specificity scores. The topics cover diverse situations ranging from facing others~(``honesty,'' ``help''), parents~(``love''), ourselves~(``self-help,'' ``self-discovery'') to facing difficulties~(``think'') and danger~(``safety'').
%standards for good or bad behavior or character: ranging from honesty to help, love and strength. %illustrating the diversity of \textsc{storal}. 
And examples of the same topic present related semantics to some extent, such as ``being honest'' and ``not believing liars'' for the first topic in \textsc{storal-en}. We also show the analysis of high-frequency words of stories and morals in Section~\ref{high-freq} and discussion about the commonsense and discourse relations in stories in Section~\ref{discuss}.

\subsection{Dataset Statistics of \textsc{storal}}  %We randomly split the labeled data for training/validation/testing of the generation models by 8:1:1 and 3:1:1 for \textsc{storal-zh} and \textsc{storal-en}, respecitvely. %We do not annotate the left unlabeled data due to the resource limitation. Instead, 
Table~\ref{tab:stat_storal} shows the statistics of \textsc{storal}.
We regard the unlabeled data which contain entangled stories and morals as an in-domain resource for research on unsupervised or semi-supervised learning for the proposed tasks. And the data are also suitable for learning to generate morals stories where the morals are weaved naturally into the story plots.  

\iffalse
In addition, to measure word overlap between morals and stories, we compute novelty-$n$, the average ratio of novel $n$-grams in a moral that do not appear in the corresponding story. We find that the novelty-1/2 score is larger than 54\%/93\% %and novelty-$3$ scores are larger than 98\% 
for both \textsc{storal-zh} and \textsc{storal-en}. The novelty-1/2 score of XSUM, the popular dataset for abstractive summarization, is 36\%/83\%~\cite{narayan2018don}. This result indicates that the \textsc{st2mo} task may be more abstractive than abstractive summarization.%. In contrast,  further indicating the difference between our and abstractive summarization. %low lexical similarity between morals and stories.
% zh: 58.07 95.73 99.18 99.68
% en: 52.24 93.94 98.62 99.57
\fi
\begin{table*}[!t]
\scriptsize
    \centering
    \begin{tabular}{l|c|ccc|ccc|cccc}
    \toprule
    
    \multirow{3}{*}{\textbf{Datasets}}& \multicolumn{7}{c|}{\textbf{Labeled Data}}&\multicolumn{3}{c}{\textbf{Unlabeled Data}}\\
    \cline{2-12}
    &\multirow{2}{*}{\textbf{\# Examples}}&\multicolumn{3}{c|}{\textbf{Stories}}&\multicolumn{3}{c|}{\textbf{Morals}}&\multirow{2}{*}{\textbf{\# Examples}}&\multirow{2}{*}{\textbf{\# Word}}&\multirow{2}{*}{\textbf{\# Sent}}&\multirow{2}{*}{\textbf{Vocab}}\\
    \cline{3-8}
    &&\textbf{\# Word} & \textbf{\# Sent} & \textbf{Vocab}&\textbf{\# Word} & \textbf{\# Sent} & \textbf{Vocab}&\\
    \hline
    \textbf{\textsc{storal-zh}}&4,209&321.75&17.62&63,493&25.09&1.35&10,522&14,988&487.00&26.12&147,805\\
    
    \textbf{\textsc{storal-en}}&1,779&302.33&17.71&15,873&19.77&1.45&3,384&819&614.55&38.05&20,853\\

   \bottomrule
    \end{tabular}
    \caption{\textsc{storal} statistics. We use Jieba\footnotemark/NLTK~\cite{loper2002nltk} for word tokenization of \textsc{storal-zh}/ \textsc{storal-en}. \textbf{\# Word} / \textbf{\# Sent} is the average number of words/sentences. \textbf{Vocab} is the vocabulary size.}
    %The abbreviation \textbf{Sent}/\textbf{Vocab} is short for sentence/vocabulary size.}%The split of dataset is based on the original STORIUM dataset, thus the validation set and the test set have different number of stories.}
    \label{tab:stat_storal}
\end{table*}

\begin{table*}[!t]
\scriptsize
    \centering
    \begin{tabular}{l|p{66pt}|p{187pt}|c|c}
    \toprule
    \multirow{2}{*}{\textbf{Tasks}}&\multirow{2}{*}{\textbf{Abilities}}&\multirow{2}{*}{\textbf{Inputs \& Outputs}}&\multicolumn{1}{c|}{\textbf{\textsc{storal-zh}}}&\multicolumn{1}{c}{\textbf{\textsc{storal-en}}}\\
    &&&\textbf{|Train|} \textbf{|Val|} \textbf{|Test|}&\textbf{|Train|} \textbf{|Val|} \textbf{|Test|}\\
    \midrule
    %\midrule
    %\multicolumn{2}{c}{\textbf{\textsc{moCpt}}}\\
    %\midrule
    \textbf{\textsc{moCpt}}&Concept Understanding&Given a story and five candidate morals, choosing the correct moral.&3,368 / 420 / 421 & 1,068 / 355 / 356\\
    \midrule
    \textbf{\textsc{moPref}}&
    Preference Alignment&Given a story and two candidate morals, choosing the correct moral.&3,276 / 410 / 411& 988 / 344 / 339\\
    \midrule
    \textbf{\textsc{st2mo}}&
    Moral Generation&Given a story, generating a moral which is character-independent and generally applicable. %and does not depend on any specific characters in the story.
    &3,368 / 420 / 421 & 1,068 / 355 / 356\\
    \midrule
    \textbf{\textsc{mo2st}}&
    Story Generation&Given a moral and a story beginning and outline, %. as an out-of-order set of phrases about characters and events
    generating a story which has a coherent plot and convinces readers of the moral.&3,368 / 420 / 421 & 1,068 / 355 / 356\\
    \bottomrule
    \end{tabular}
    \caption{Description of the proposed tasks about the abilities they investigate, inputs and outputs, and the data sizes.} %for training, validation and testing. }%The number before/after the slash indicates the size for \textsc{storal-zh}/\textsc{storal-en}, respectively.}
    \label{tab:stat}
\end{table*}

\subsection{Task-Specific Dataset Construction}
Based on \textsc{storal}, we build task-specific datasets for our understanding tasks~(\textsc{moCpt} and \textsc{moPref}) and generation tasks~(\textsc{st2mo} and \textsc{mo2st}). 
We randomly split the labeled data in \textsc{storal-zh} and \textsc{storal-en} for training/validation/testing by 8:1:1 and 3:1:1, respectively. Table~\ref{tab:stat} shows the task descriptions and data sizes.
\paragraph{\textsc{moCpt}}{} It requires selecting the correct moral from five candidates given a story. We constructed the dataset by taking the original moral as the correct candidate and four negatively sampled morals as incorrect candidates for each example. To avoid more than one plausible candidate, we ensured that the negative morals are assigned to different topics from the original one by the LDA model~(Section~\ref{topic_ana}). In this way, \textsc{moCpt} can effectively test the ability to distinguish different concepts.

\paragraph{\textsc{moPref}}{} It requires selecting the correct moral from two candidates. Its difference from \textsc{moCpt} is that we created the incorrect candidate by substituting one random token in the original moral to its antonym. %, or adding or deleting negation words. 
For example, the moral ``unity is strength'' can be transformed to ``unity is weakness''. We perform the transformation using a rule-based method~\cite{ribeiro-etal-2020-beyond}. Because there exist examples where no words have antonyms, the number of examples for \textsc{moPref} are a little fewer than \textsc{moCpt}. %because the automatic method may fail to transform. 
\textsc{moPref} will serve for testing the ability  to capture the value preference of stories.

\paragraph{\textsc{st2mo}}{} It requires generating the moral of  a given story. We regard the original story as input and the original moral as target output.

\paragraph{\textsc{mo2st}}{} It requires generating a story to convey a given moral. Unfortunately, %it is still difficult for existing generation models to expand a reasonable plot conditioned only on limited and abstract information~\cite{guan2020knowledge}. In addition, 
automatic evaluation for open-ended story generation is still highly challenging due to the notorious one-to-many issue~\cite{zhao2017learning}: There may be multiple plausible stories with the same moral. For example, the moral in Table~\ref{tab:example_story} can also be conveyed by another story: {``bees unite to build their beehives.''}  %other stories such as 
 %\textit{``some scrawny tigers and lions unite to defeat the strong cows''}. 
Such openness makes automatic metrics %brings a huge challenge for 
unreliable for quality evaluation~\cite{guan2020union}.

To alleviate this issue, we extract the first sentence and an outline from a target story, and pair them with the moral as input for generating the story. We follow \citet{rashkin2020plotmachines} to extract a set of at most eight phrases from a story through RAKE~\cite{rose2010automatic} as the outline. We set the maximum number of words in each phrase to eight. We also filtered those phrases that are substrings of others. For example, the outline for the story in Table~\ref{tab:example_story} is \{{``lions,''} {``friends fought,''} {``good friends,''} {``grazed,''} {``perfect opportunity''}\}. Finally, for \textsc{storal-zh}/\textsc{storal-en}, the average number of phrases for each example is 7.5/6.8 and the average number of words in each phrase is 2.87/2.44, respectively. %The outline can %serve as discourse-level guidance for the subsequent generation, which 
%not only narrow down the set of plausible stories but also serve as discourse-level guidance to enforce models to rearrange the events reasonably %and generate a story with a good global discourse structure 
%rather than focus on modeling local coherence. %Then we concatenate moral, keyphrases with random order and the first sentence of story as input to generate the remaining story.

\footnotetext[1]{\url{https://github.com/fxsjy/jieba}}

\section{Retrieval Augmentation}
\begin{figure}[!t]
\includegraphics[width=\linewidth]{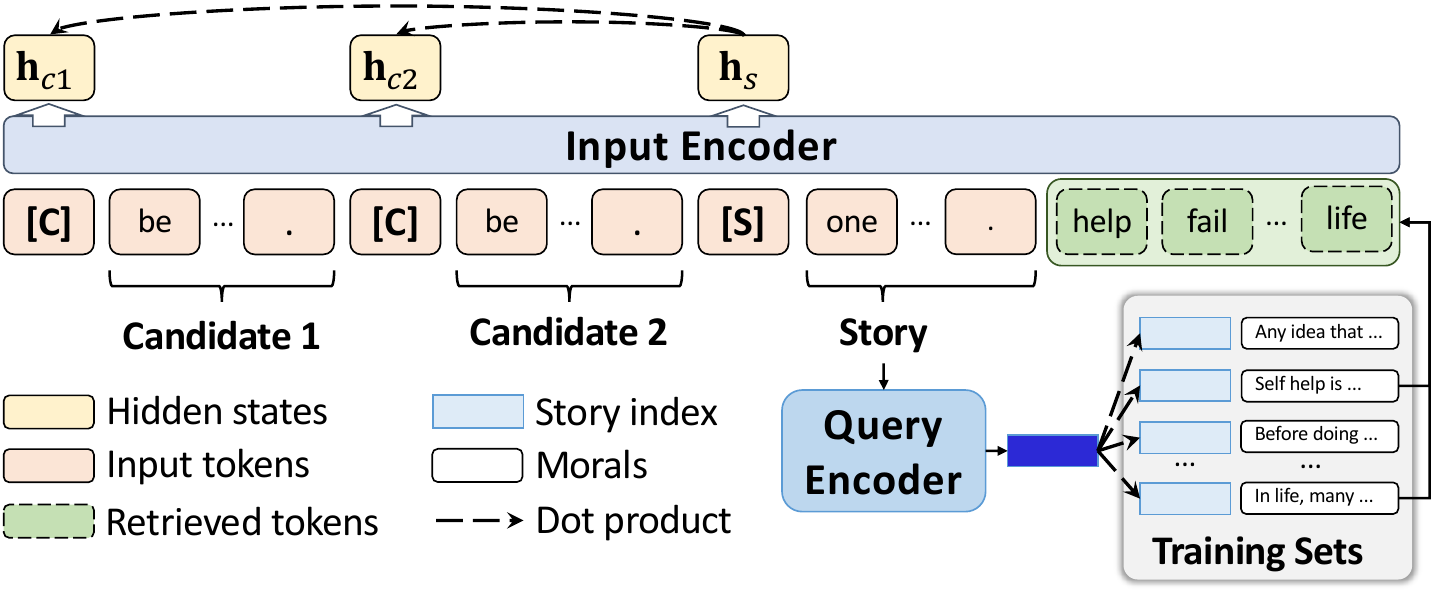}% 1\linewidth
  \caption{Model overview for the \textsc{moPref} task.}%, the discourse structure of the story~(Bottom Left) and the association between the story and the moral~(Bottom Right).}
  \label{fig:our_model}
\end{figure}

A critical challenge for tackling the proposed tasks is the sparsity of morals and events makes it difficult to learn relations between them. Prior studies have shown that retrieval improves performance towards infrequent data points across various tasks such as open-domain question answering~\cite{chen-etal-2017-reading} and text classification~\cite{lin2021bertgcn}. We present a retrieval-augmented algorithm that exploits the moral-event relations in training sets. We illustrate our model for the \textsc{moPref} task in Figure~\ref{fig:our_model}. Our models for other tasks are similar.

For both \textsc{moCpt} and \textsc{moPref}, we encode the story and candidates using an input encoder, and then predict a probability distribution over the candidates by normalizing the dot-product scores between the representations of the story and each candidate. We optimize the model by minimizing the cross-entropy loss. We insert special tokens \texttt{[S]} and \texttt{[C]} before the story and each candidate, respectively, and take the corresponding hidden states as their representations. %To alleviate the data sparsity issue, 
Furthermore, we propose to retrieve related concepts from the training set using the input story. We encode the story using a query encoder, then take the output as the query to retrieve $m$ most related stories based on a story index, i.e., a set of dense vectors as the representations of stories in the training set.  We adopt BERT~\cite{devlin2018bert} followed by a mean-pooling layer to build the query encoder and story index, which are frozen in the training stage. Finally, we extract the nouns, verbs, adjectives and adverbs from the morals of the top-$m$ stories and lemmatize them as the retrieved concepts. We feed the concepts together with the original input to the input encoder. For example, the retrieved concepts for the story in Table~\ref{tab:example_story} include ``support'' and ``strength'', which may serve as additional guidance for models' prediction.

The retrieval-augmented algorithm can easily adapt to the generation tasks. For \textsc{st2mo}, we take the input story paired with the retrieved concepts into the encoder and then generate the output using the decoder. And for \textsc{mo2st}, we use the input moral as the query to retrieve top-$m$ stories, and regard their outlines as the retrieved additional information to guide the subsequent story generation.

%Prior work has shown that retrieval improves performance across a variety of tasks such as open-domain question answering

\section{Experiments}
%In this section, we evaluate existing models on the proposed tasks, and propose a retrieval based method for . %analyze their strengths and weaknesses with extensive experiments. %Formally, the general conditional language generation tasks can be defined as follows: given an input, the model should output a natural language text adhering to the input. For \textsc{st2mo}, the input is a story and the output is the corresponding moral. 
%We follow \citet{rashkin2020plotmachines} to define the outline as a set of out-of-order phrases that describe key characters and events to appear in a story, and extract the phrases from a story automatically using the RAKE algorithm~\cite{rose2010automatic}.

%s2s: 50M 47M
%fusion: 100M 94M 
%gpt2: ... 124M
%t5 220M 220M
\subsection{Evaluated Models}
We evaluated the following baselines for the understanding tasks: %\textit{Vanilla Transformer}~\cite{vaswani2017attention}, which consists of three transformer blocks; 
\textit{{BERT}}~\cite{devlin2018bert}, \textit{{RoBERTa}}~\cite{liu2019roberta} and \textit{{T5}}~\cite{raffel2020exploring}. %, which achieve state-of-the-art performance on various tasks. 
When evaluating T5, we feed the input to both the encoder and decoder of T5 and optimize the model using the cross-entropy loss. To investigate potential biases of the proposed datasets, we added a baseline called \textit{BERT w/o story}, which is fine-tuned to make prediction without taking the story as input. For the generation tasks, we evaluated \textit{ConvS2S}~\cite{gehring2017convolutional}, \textit{Fusion}~\cite{fan2018hierarchical}, \textit{GPT2}~\cite{radford2019language} and \textit{T5}, 
%\textbf{(1) Non-pretrained models:} including \textit{ConvS2S}~\cite{gehring2017convolutional} and the \textit{Fusion} model~\cite{fan2018hierarchical}. \textbf{(2) General pretrained models:} including \textit{GPT2}~\cite{radford2019language} %, BART~\cite{bart} 
%and \textit{T5}~\cite{raffel2020exploring}, which achieve state-of-the-art performance on various generation tasks. 
which are trained or fine-tuned with the standard language modeling objective. Moreover, we also evaluate a task-specific model %including %{KG-GPT2}~\cite{guan2020knowledge} which post-trains GPT2 on external knowledge bases, and 
{\textit{PlotMachines}}~({PM} for short)~\cite{rashkin2020plotmachines}, which is proposed for tackling outline-conditioned generation by tracking the dynamic plot states. We use GPT2 as the backbone model of PM.

We also design models to test the adaption of the unlabeled data of \textsc{storal} to the proposed tasks. Specifically, we first post-train RoBERTa and T5 on the unlabeled data with their original pretraining objectives, respectively (i.e., masked language model and text infilling) and then fine-tune them on the labeled data for the downstream tasks~\cite{gururangan2020don}. We call the baselines \textit{RoBERTa-Post} and \textit{{T5-Post}}. We perform our retrieval-augmented algorithm based on the post-trained models, called \textit{RA-RoBERTa} and \textit{RA-T5}, respectively.

\subsection{Experiment Settings}
%the non-pretrained models based on the tools provided by the original papers 
We implement the pretrained models based on the codes and pretrained checkpoints of HuggingFace's Transformers~\cite{wolf2020transformers}. We use LongLM$_{\rm base}$~\cite{guan2021lot} as the T5 model for experiments on \textsc{storal-zh}, %\footnote{\url{https://github.com/huggingface/ transformers}}. 
and set all pretrained models to the base version due to limited computational resources. As for the hyper-parameters, we set the batch size to 16, the maximum sequence length to 1,024, the learning rate to 3e-5, $m$ to 10 for our retrieval-augmented model. %The batch size is set to 16 and the maximum sequence length to 1,024. 
%We use the Adam optimizer with an initial learning rate of 3e-5. 
We generate outputs using top-$k$ sampling~\cite{fan2018hierarchical} with $k=40$ and a softmax temperature of 0.7~\cite{goodfellow2016deep}. %to balance the trade-off between diversity and fluency.
We show more details  in Section~\ref{imple}.

\subsection{Automatic Evaluation}
\paragraph{Evaluation Metrics} We adopt accuracy to evaluate the understanding tasks. For generation tasks, we do not use perplexity since perplexity scores are not comparable among models with different vocabularies. We adopt the following metrics for automatic evaluation: %\textbf{(1) Perplexity~(PPL)}.
\textbf{(1) BLEU (B-$n$):} It is used to measure $n$-gram overlaps~($n=1,2$) between generated and ground-truth texts~\cite{papineni2002bleu}.  \textbf{(2) BERTScore-F1~(BS):} It is used to measure the semantic similarity between generated and ground-truth texts~\cite{zhang2019bertscore}. \textbf{(3) Repetition~(R-$n$):} It calculates the ratio of texts that repeat at least one $n$-gram in all generated texts~\cite{shao-etal-2019-long}. \textbf{(4) Distinct (D-$n$):} It measure the diversity using the percentage of distinct $n$-grams to all $n$-grams in generated texts~\cite{li2016diversity}. For both R-$n$ and D-$n$, we set $n=2$ for \textsc{st2mo} and $n=4$ for \textsc{mo2st} considering the much shorter length of morals than stories. %\textbf{(5) Novelty (Novel-4):} The metric computes the average ratio of novel $n$-grams in the moral that do not appear in the story. For \textsc{st2mo}, we compute novelty using the given story and the generated moral, and it is the exact opposite for \textsc{mo2st}. A higher novelty indicates a better ability to understand and generate abstract texts in contrast to simply copying words.
Besides, we also report the average number of generated words~(\textbf{Len}).

We also adopt the following metrics for automatic evaluation of \textsc{mo2st}: \textbf{(1) Coverage~(Cov):} It computes Rouge-L recall~\cite{lin-2004-rouge} between generated stories and phrases in the corresponding outlines. %as the coverage score. 
A higher score means the generated stories cover more phrases in the given outlines. \textbf{(2) Order (Ord):} It measures the disparity between the positional orders of given phrases in the ground truth and generated story using the percentage of inversions in the generated story~\cite{guan2021lot}. An inversion is a position pair of two phrases that is out of the ground-truth order. Higher order scores mean that the stories arrange the outline more reasonably. %And we use the position of the longest common subsequence between a story and a phrase as the position of the phrase in the story.
%Considering that an input phrase does not always appear in the generated story, we regard all the position pairs of such phrase and others as inversions.
In Section~\ref{auto_cons}, we also construct a learnable automatic metric to measure the faithfulness between morals and stories.
%Note that we do not adopt perplexity for automatic evaluation, the standard language modeling metric, since the perplexity scores are not comparable between models with different vocabularies.

\paragraph{Results} Table~\ref{tab:select_acc_test} and \ref{tab:auto_eva} show the results on the understanding and generation tasks, respectively. To get the human performance on \textsc{moCpt} and \textsc{moPref}, we sampled 100 examples from the test set and recruited three annotators with good Chinese or English language proficiency to complete these tasks. We made final decisions among the annotators through major voting. The annotation results show an almost perfect agreement with Fleiss's $\kappa>0.85$~\cite{Fleiss1971Measuring}.

\begin{table}[!t]
\scriptsize
    \centering
    \begin{tabular}{l|c|ll|ll}
    \toprule
    \multirow{2}{*}{\textbf{Models}}&\multirow{2}{*}{\textbf{\# P}}&\multicolumn{2}{c|}{\textbf{\textsc{moCpt}}}&\multicolumn{2}{c}{\textbf{\textsc{moPref}}}\\
    &&\textbf{\textsc{zh}}&\textbf{\textsc{\textbf{en}}}&\textbf{\textsc{zh}}&\textbf{\textsc{en}}\\
    \midrule
    %\textbf{\# Examples}&N/A&421&356&415&328\\
    %\midrule
    \textbf{Random}&N/A&20.19&20.22&50.12&50.00 \\
    \textbf{BERT w/o Story}&110M&23.52&22.47&71.81&72.57\\
    \midrule
    %\textbf{Tfmr}&40M&21.85&18.82&51.34&50.91\\
    \textbf{BERT}&110M&59.62&51.97&82.97&79.35\\
    \textbf{RoBERTa}&110M&62.71&54.78&\underline{89.54}&81.12\\
    \textbf{RoBERTa-Post}&110M&64.61&51.40&87.59&{81.42}\\
    \textbf{T5}&220M&69.60&58.99&82.00&76.99\\
    \textbf{T5-Post}&220M&\underline{70.07}&\underline{62.64}&81.75&77.29\\
    \midrule
    \textbf{RA-RoBERTa}&110M&65.08&60.96&\textbf{90.02}&\underline{81.71}\\
    \textbf{RA-T5}&220M&\textbf{72.68}$^{*}$&\textbf{67.42}${^{**}}$&82.97&\textbf{82.60}\\
    \midrule
    \textbf{Human}&N/A&\textit{95.00}&\textit{96.00}&\textit{98.00}&\textit{99.00}\\
    %97: kappa=0.24
    \bottomrule
    \end{tabular}
    \caption{Accuracy~(\%) for \textsc{moCpt} and \textsc{moPref}. \# P is the number of parameters. The best performance is highlighted in \textbf{bold} and the second best is underlined. The scores marked with $^{*}$ and $^{**}$ of RA model mean it outperforms the best baseline significantly with p-value$<$0.1 and p-value$<$0.05~(sign test), respectively.}
    \label{tab:select_acc_test}
\end{table}

\begin{table*}[!t]
\scriptsize
    \centering
    \begin{tabular}{l|l|llllll|llllll}
    \toprule
    \multirow{2}{*}{\textbf{Models}}&\multirow{2}{*}{\textbf{{\# P}}}&\multicolumn{6}{c|}{\textbf{Dataset: \textsc{storal-zh}}}&\multicolumn{6}{c}{\textbf{Dataset: \textsc{storal-en}}}\\
    %\multicolumn{13}{c}{\textbf{Task: \textsc{st2mo}}}\\
    %\textbf{Metrics}
    &&\textbf{B-1}$\uparrow$&\textbf{B-2}$\uparrow$&\textbf{BS}$\uparrow$&\textbf{R-2}$\downarrow$&\textbf{D-2}$\uparrow$&\textbf{Len}&\textbf{B-1}$\uparrow$&\textbf{B-2}$\uparrow$&\textbf{BS}$\uparrow$&\textbf{R-2}$\downarrow$&\textbf{D-2}$\uparrow$&\textbf{Len}\\    
    \midrule
    \textbf{ConvS2S}&50M&14.31&1.86&56.71&\textbf{26.60}&43.67&19.31&9.69&0.93&82.57&\underline{6.46}&47.35&11.75\\
    \textbf{Fusion}&100M&14.78&2.23&56.90&\underline{27.55}&41.21&21.96&9.87&0.82&82.68&\textbf{6.18}&43.59&13.15\\
    \midrule
    \textbf{GPT2}&124M&14.54&2.16&60.75&35.39&\underline{48.22}&20.72&10.98&1.24&79.39&20.22&\underline{60.36}&16.19\\
    %\textbf{BART}&N/A&N/A&N/A&N/A&N/A&N/A\\
    \textbf{T5}&220M&\underline{18.19}&{3.60}&\underline{61.61}&76.48&44.84&29.06&{13.31}&\underline{2.26}&\underline{85.89}&33.15&{58.73}&19.39\\
    \textbf{T5-Post}&220M&{17.98}&\textbf{3.91}&{61.52}&69.12&\textbf{51.97}&29.14&\underline{13.83}&{2.11}&{85.85}&34.83&57.12&18.49\\
    \midrule
    \textbf{RA-T5}&220M&\textbf{18.32}&\underline{3.64}&\textbf{61.93}$^{**}$&70.78&48.14&29.44&\textbf{14.59}&\textbf{2.61}&\textbf{86.16}$^{**}$&31.46&\textbf{60.61}&18.54\\
    \midrule
    %\textbf{KG-GPT2} (124M)&N/A&N/A&N/A&N/A&N/A&N/A\\
    %\midrule
    \textbf{\textit{Truth Morals}}&\textit{N/A}&\textit{N/A}&\textit{N/A}&\textit{N/A}&\textit{29.22}&\textit{73.70}&\textit{25.09}&\textit{N/A}&\textit{N/A}&\textit{N/A}&\textit{16.85}&\textit{73.95}&20.41\\
    \bottomrule
    \end{tabular}
    \tiny
    \begin{tabular}{l|cccccccc|cccccccc}
    \toprule
    \multirow{2}{*}{\textbf{Models}}&\multicolumn{8}{c|}{\textbf{Dataset: \textsc{storal-zh}}}&\multicolumn{8}{c}{\textbf{Dataset: \textsc{storal-en}}}\\    
    
    &\textbf{B-1}$\uparrow$&\textbf{B-2}$\uparrow$&\textbf{BS}$\uparrow$&\textbf{R-4}$\downarrow$&\textbf{D-4}$\uparrow$%&\textbf{Faith}$\uparrow$
    &\textbf{Cov}$\uparrow$&\textbf{Ord}$\uparrow$&\textbf{Len}&\textbf{B-1}$\uparrow$&\textbf{B-2}$\uparrow$&\textbf{BS}$\uparrow$&\textbf{R-4}$\downarrow$&\textbf{D-4}$\uparrow$%&\textbf{Faith}$\uparrow$
    &\textbf{Cov}$\uparrow$&\textbf{Ord}$\uparrow$&\textbf{Len}\\        
    \midrule
    \textbf{ConvS2S}&15.57&6.43&60.00&\underline{75.30}&\underline{78.41}&%33.44&
    21.61&33.03&150&16.25&6.38&79.27&\underline{61.85}&\textbf{80.29}%&35.59
    &6.46&41.88&122\\
    \textbf{Fusion}&15.53&6.45&60.06&\textbf{74.11}&\textbf{80.51}%&38.81
    &22.86&33.33&148&17.17&6.82&79.52&\textbf{61.24}&75.79&%35.16&
    7.27&43.07&137\\
    \midrule
    \textbf{GPT2}&14.91&6.48&63.32&91.45&58.67&%50.49&
    48.57&51.58&282&25.83&12.91&{83.25}&84.27&74.63%&64.90
    &45.18&59.95&247\\
    \textbf{\textsc{PM}}&15.82&7.04&63.58&90.97&57.33&%52.41&
    50.51&52.35&280&26.34&13.92&81.63&80.90&72.64&%62.53
    47.07&60.31&264\\
    %\textbf{BART}&N/A&N/A&N/A&N/A&N/A&N/A\\
    \textbf{T5}&{17.74}&{9.44}&\underline{65.89}&91.69&{61.76}&%\underline{56.66}&
    {58.18}&{56.11}&166&{30.56}&{16.75}&79.89&90.17&\underline{77.53}&%{71.63}&
    {74.21}&{63.45}&283\\
    \textbf{T5-Post}&\underline{18.42}&\underline{9.77}&{65.63}&94.54&58.13&%\textbf{58.58}&
    \underline{60.11}&\underline{56.96}&176&\underline{32.36}&\underline{18.04}&\underline{83.80}&94.10&77.27&%\underline{75.67}&
    \underline{76.09}&\underline{64.33}&281\\
    %\textbf{KG-GPT2}~(124M)&N/A&N/A&N/A&N/A&N/A&N/A\
    \midrule
    \multirow{2}{*}{\textbf{RA-T5}}&\textbf{23.36}&\textbf{12.98}&\textbf{67.37}&95.72&59.49&\textbf{69.24}&\textbf{60.44}&241&\textbf{32.46}&\textbf{18.31}&\textbf{84.07}&{92.42}&{76.74}&%74.92
    \textbf{80.21}&\textbf{66.10}&253\\
    &**&**&**&&&**&**&&&*&**&&&**&**\\
    \midrule
    \textbf{\textit{Truth}}&\textit{N/A}&\textit{N/A}&\textit{N/A}&\textit{55.34}&\textit{96.06}&%\textit{77.49}&
    \textit{100.00}&\textit{100.00}&324&\textit{N/A}&\textit{N/A}&\textit{N/A}&\textit{58.71}&\textit{95.09}&%\textit{80.03}&
    \textit{100.00}&\textit{100.00}&281\\
    \bottomrule
    \end{tabular}
    \caption{Automatic evaluation results for \textsc{st2mo}~(Top) and \textsc{mo2st}~(Bottom). $\downarrow$ / $\uparrow$ means the lower/higher the better. All scores except \textit{Len} are multiplied by 100. %Other settings are the same as Table~\ref{tab:select_acc_test}.%\#Params means the number of parameters. $\downarrow$ / $\uparrow$ means the lower/higher the better. %We present the results of perplexity just as a reference for future work. 
    %N/A means that there are not public Chinese models currently. 
    %All scores are multiplied by 100. 
    The best result is in \textbf{bold} and the second best is underlined. The scores marked with $^{*}$ and $^{**}$ of RA-T5 mean it outperforms the best baseline significantly with p-value$<$0.1 and p-value$<$0.05~(sign test), respectively.
    }%The scores marked with  means the model outperforms the previous one significantly with $p<0.01$~(sign test). Note that we do not conduct the significance test }
    \label{tab:auto_eva}
\end{table*}

%We show the automatic evaluation results on the test set of \textsc{storal} in Table~\ref{tab:auto_eva}/\ref{} for \textsc{st2mo}/\textsc{mo2st}, respectively. 
We summarize the results on the understanding tasks as follows: \textbf{(1)} The \textsc{moPref} datasets suffer from innate biases as indicated by the high accuracy of BERT w/o story. Such biases may result from the noise introduced by the automatic construction technique, i.e., antonym substitution. And models may learn patterns of %since humans can easily distinguish the 
good behavior~(e.g., ``unity'' is good and ``disunity'' is bad in general) and make predictions easily without depending on stories. However, \textsc{moPref} is still meaningful as an evaluation task since BERT can achieve much better accuracy when taking stories as input. And we experiment using manually constructed examples for evaluating preference alignment in the appendix. \textbf{(2)} T5 performs better than RoBERTa on \textsc{moCpt} but worse on \textsc{moPref}, indicating T5 may not be good at capturing value preferences. \textbf{(3)} Post-training on the unlabeled data~(i.e., RoBERTa-Post and T5-Post) does not always bring improvement on both tasks, suggesting that it is necessary to develop a better way to exploit these data in future work. \textbf{(4)} Retrieving additional concepts improves models' performance effectively, particularly for the \textsc{moCpt} task on \textsc{storal-en}. However, there is still a big gap between our models and human performance.

As for the generation tasks, we draw the following conclusions: \textbf{(1)} Almost all pretrained models achieve better lexical and semantic similarity with ground-truth texts than non-pretrained models,
%Compared with non-pretrained models, almost all the pretrained models achieve better lexical and semantic similarity with ground-truth texts, 
as indicated by higher BLEU and BERTScore values. %The superiority is more significant when generating stories than morals. 
\textbf{(2)} %As prior studies observed~\cite{DBLP:conf/emnlp/XuPSPFAC20},
Non-pretrained models have less repetition than pretrained ones, and repeat even less than the ground-truth texts when generating morals. It may be because %small models are better at capturing short-term dependencies. Furthermore, when generating stories, we observe that 
non-pretrained models generate shorter sequences than pretrained models despite the same decoding algorithm, which also accounts for the higher distinct scores of the non-pretrained models on the \textsc{mo2st} task. %leads to less repetition and higher distinct. %the non-pretrained models generate more diverse $n$-grams than pretrained models for \textsc{mo2st}, which 
%which is because also verifies the difficulty for non-pretrained models to capture the subtle dependencies between inputs and outputs. 
%Although pretraining can improve the situation, there still exists a wide gap with ground-truth texts in the generation diversity. 
\textbf{(3)} When generating stories, T5-Post can cover more input phrases and arrange them in a correct order than other baselines, as indicated by higher coverage and order scores. \textbf{(4)} %Dynamically tracking the plot states (i.e., PM) or post-training (i.e., T5-Post) do not bring significant improvement on almost all metrics, 
Retrieval augmentation can improve the generation similarity with the ground-truth texts on both tasks and improve the coverage and order scores on \textsc{st2mo} significantly compared with T5-Post.

%for the \textsc{mo2st} task, dynamically tracking the plot states with a memory network can improve the BLEU but 

%we notice that post-training T5 on the unlabeled data can further improve the consistency, suggesting the general benefit of the data. And 

\begin{table}[!t]
\scriptsize
    \centering
    \begin{tabular}{c|c|l|c|c|c}
    \toprule
    \textbf{Data}&\textbf{Task}&\textbf{Model}&\textbf{Flu ($\kappa$)}&\textbf{Cohe~($\kappa$)}&\textbf{Faith~($\kappa$)}\\
    \midrule
    %\midrule
    %\multicolumn{5}{c}{\textbf{Dataset: \textsc{storal-zh}}}\\
    %\midrule
    \multirow{8}{*}{\rotatebox{90}{\textbf{\textsc{storal-zh}}}}&\multirow{4}{*}{\rotatebox{90}{\textbf{\textsc{st2mo}}}}&\textbf{Fusion}&0.24 (0.31) &0.22 (0.37)&0.08 (0.72) \\
    &&\textbf{T5}&0.75 (0.40)&0.61 (0.38)&0.31 (0.32) \\
    &&\textbf{RA-T5}&\textbf{0.85}~(0.65)&\textbf{0.63}~(0.26)&\textbf{0.36}~(0.27)\\
    \cline{3-6}
    &&\textbf{{\textit{Truth}}}&\textit{1.00 } (1.00)&\textit{0.99} (0.96)&\textit{0.86} (0.57)\\
    %\midrule
    \cline{2-6}
    &\multirow{4}{*}{\rotatebox{90}{\textbf{\textsc{mo2st}}}}&\textbf{Fusion}&0.25 (0.39)&0.11 (0.61)& 0.02 (0.93)\\
    &&\textbf{T5} &0.38 (0.40)&0.24 (0.37)&0.05 (0.81)\\
    &&\textbf{RA-T5}&\textbf{0.45}~(0.29)&\textbf{0.34}~(0.25)&\textbf{0.11}~(0.72)\\
    \cline{3-6}
    &&\textbf{{\textit{Truth}}}&\textit{0.98} (0.93)&\textit{1.00} (1.00)&\textit{0.96} (0.84)\\
    %\midrule
    \midrule
    %\multicolumn{5}{c}{\textbf{Dataset: \textsc{storal-en}}}\\
    %\midrule
    
    \multirow{8}{*}{\rotatebox{90}{\textbf{\textsc{storal-en}}}}&\multirow{4}{*}{\rotatebox{90}{\textbf{\textsc{st2mo}}}}&\textbf{Fusion}&0.32 (0.39) &0.26 (0.35)&0.24 (0.41) \\
    &&\textbf{T5}&0.76 (0.35)&0.74 (0.27)&0.55 (0.33) \\
    &&\textbf{RA-T5}&\textbf{0.81}~(0.51)&\textbf{0.79}~(0.40)&\textbf{0.67}~(0.37)\\
    \cline{3-6}
    &&\textbf{{\textit{Truth}}}&\textit{0.94 }(0.80)&\textit{0.94} (0.77)&\textit{0.88} (0.56)\\
    %\midrule
    \cline{2-6}
    &\multirow{4}{*}{\rotatebox{90}{\textbf{\textsc{mo2st}}}}&\textbf{Fusion}&0.47 (0.43)&0.40 (0.47)& 0.37 (0.45)\\
    &&\textbf{T5} &0.56 (0.35)&0.48 (0.37)&0.49 (0.39)\\
    &&\textbf{RA-T5}&\textbf{0.58}~(0.28)&\textbf{0.51}~(0.31)&\textbf{0.57}~(0.31)\\
    \cline{3-6}
    &&\textbf{{\textit{Truth}}}&\textit{0.95} (0.69)&\textit{0.98} (0.69)&\textit{0.93} (0.53)\\    
    \bottomrule    
    \end{tabular}
    \caption{Manual evaluation results for \textsc{st2mo} and \textsc{mo2st}. Flu, {{Cohe}} and {Faith} mean \textit{fluency}, \textit{coherence} and \textit{moral faithfulness}, respectively. The best results are highlighted in \textbf{bold}. All results show fair or moderate inter-annotator agreement measured by Fleiss' $\kappa$~\cite{Fleiss1971Measuring}.}
    \label{tab:man_eva}
\end{table}

\subsection{Manual Evaluation}
%Since existing automatic metrics are not reliable to evaluate language generation, 
On the generation tasks, we conducted a Likert-scale based manual evaluation to measure the gap between existing models and humans. For \textsc{stoal-zh}, we hired three graduate students~(native Chinese speakers) as annotators. We conducted evaluation on \textsc{storal-en} using Amazon Mechanical Turk~(AMT). For both tasks, we randomly sampled 100 examples from the test set, and obtained 300 generated texts from Fusion, T5 and RA-T5. For each text %~(either written by humans or generated by a  model) 
%along with the input, 
we require three annotators to rate its quality along with the input using a binary score in three following aspects: (1) \textit{linguistic fluency}: correctness in grammaticality; %~(intra-sentence linguistic quality and grammatical correctness), 
(2) \textit{coherence}: reasonable relations between sentences regarding %~(inter-sentence 
relatedness, causality and temporal orders; 
and (3) \textit{moral faithfulness}: exhibition of a faithful moral to the input. Three aspects are independently evaluated. We decided the final score of a text through majority voting. The annotation instruction is shown in Section~\ref{instruc}.

Table~\ref{tab:man_eva} shows the manual evaluation results. We show $p$-values of the results %when comparing each pair of the ground truth and three generation models 
in Section~\ref{signi}. 
For \textsc{st2mo}, T5 achieves a substantial improvement compared with Fusion~($p<0.01$), %suggesting the benefit of large-scale pretraining.
and our model further outperforms T5. The superiority becomes less significant for \textsc{mo2st}. However, the big gap between these models and humans, particularly in terms of faithfulness, proves both tasks challenging for existing models. Furthermore, we evaluate whether machines can capture the value preference of a story using manually constructed examples. And we show error analysis and case study for the proposed tasks in Section~\ref{case_error_study}.   We believe that explicit modeling of the relations among events and abstract concepts will further promote progress on these tasks, which we regard as future work. %which is difficult to reveal by automatic metrics. 

%is still a long way for generation models to achieve human-level performance particularly in  .

%Furthermore, we also ask annotators to annotate the specific errors in the generated texts.

%\paragraph{Evaluation for \textsc{st2mo}}
%\paragraph{Evaluation for \textsc{mo2st}}

\section{Conclusion}
We present \textsc{storal}, a collection of Chinese and English moral stories. To test the ability to bridge concrete events and abstract morals, we propose new understanding and generation tasks based on \textsc{storal}, including selecting the correct moral from several candidates with different topics or opposite value preferences, concluding the moral of a story and generating a story to convey a moral. Extensive experiments prove these tasks still to be challenging for existing models. We propose a retrieval-augmented algorithm to improve performance by retrieving related concepts or events from training sets. Although it is possible to further increase the dataset size, we expect to make meaningful progress by developing better representations of commonsense and discourse relations among events and abstract concepts in future work.
\section{Acknowledgement}
This work was supported by the National Science Foundation for Distinguished Young Scholars (with No. 62125604) and the NSFC projects (Key project with No. 61936010 and regular project with No. 61876096). This work was also supported by the Guoqiang Institute of Tsinghua University, with Grant No. 2019GQG1 and 2020GQG0005. This work was also sponsored by Tsinghua-Toyota Joint Research Fund. We would also like to thank the anonymous reviewers for their invaluable suggestions and feedback.

\section{Ethics Statements and Broader Impact}
We collected \textsc{storal} from public web resources. All stories are under licenses that allow use and redistribution for research purposes. We asked commercial annotation teams to extract stories and morals from the crawled raw texts. We required the annotators to refuse the examples which violate general ethical principles~(e.g., showing discrimination for someone, containing disrespectful content, or encouraging to disturb public order, etc.). Totally, we payed more than \$7 (CNY 45) per hour on average for annotating each example in \textsc{storal}, which was far beyond the minimum hourly wage in China~(CNY 21). Furthermore,
%build OpenMEVA based on two existing public story datasets ROCStories~(ROC) and WritingPrompts~(WP), which are widely used for story generation and evaluation. 
we resorted to AMT for manual evaluation of generated and human-written texts for two proposed generation tasks. We hired three annotators and payed each annotator \$0.2 on average for annotating each example. %We did not ask about personal privacy or collect personal information of annotators in the above annotation processes.   %Finally we obtained 2,000 annotated stories~(1,000 for ROC and 1,000 for WP) and spent about \$900 (including extra \$150 for fees to AMT). 
%We admit that 
%There may be still unpredictable bias in \textsc{storal} even though we have reviewed all annotated examples. 

In this paper, we emphasize the ability to model relations between concrete events and abstract morals, which is also helpful for various scenarios such as reading comprehension~(e.g., drawing authors' viewpoints from narratives) and essay writing~(e.g., writing essays to convince readers of some arguments by presenting examples or anecdotes). \textsc{storal} provides a good start point for exploring these directions.
\bibliography{anthology,custom}

\begin{thebibliography}{56}
\expandafter\ifx\csname natexlab\endcsname\relax\def\natexlab#1{#1}\fi

\bibitem[{Akoury et~al.(2020)Akoury, Wang, Whiting, Hood, Peng, and
  Iyyer}]{akoury2020storium}
Nader Akoury, Shufan Wang, Josh Whiting, Stephen Hood, Nanyun Peng, and Mohit
  Iyyer. 2020.
\newblock \href {https://doi.org/10.18653/v1/2020.emnlp-main.525} {{STORIUM}:
  {A} {D}ataset and {E}valuation {P}latform for {M}achine-in-the-{L}oop {S}tory
  {G}eneration}.
\newblock In \emph{Proceedings of the 2020 Conference on Empirical Methods in
  Natural Language Processing (EMNLP)}, pages 6470--6484, Online. Association
  for Computational Linguistics.

\bibitem[{Blei et~al.(2003)Blei, Ng, and Jordan}]{blei2003latent}
David~M Blei, Andrew~Y Ng, and Michael~I Jordan. 2003.
\newblock Latent dirichlet allocation.
\newblock \emph{the Journal of machine Learning research}, 3:993--1022.

\bibitem[{Brahman and Chaturvedi(2020)}]{brahman2020modeling}
Faeze Brahman and Snigdha Chaturvedi. 2020.
\newblock \href {https://doi.org/10.18653/v1/2020.emnlp-main.426} {Modeling
  protagonist emotions for emotion-aware storytelling}.
\newblock In \emph{Proceedings of the 2020 Conference on Empirical Methods in
  Natural Language Processing (EMNLP)}, pages 5277--5294, Online. Association
  for Computational Linguistics.

\bibitem[{Chen et~al.(2017)Chen, Fisch, Weston, and
  Bordes}]{chen-etal-2017-reading}
Danqi Chen, Adam Fisch, Jason Weston, and Antoine Bordes. 2017.
\newblock \href {https://doi.org/10.18653/v1/P17-1171} {Reading {W}ikipedia to
  answer open-domain questions}.
\newblock In \emph{Proceedings of the 55th Annual Meeting of the Association
  for Computational Linguistics (Volume 1: Long Papers)}, pages 1870--1879,
  Vancouver, Canada. Association for Computational Linguistics.

\bibitem[{Cui et~al.(2020)Cui, Che, Liu, Qin, Wang, and
  Hu}]{cui-etal-2020-revisiting}
Yiming Cui, Wanxiang Che, Ting Liu, Bing Qin, Shijin Wang, and Guoping Hu.
  2020.
\newblock \href {https://www.aclweb.org/anthology/2020.findings-emnlp.58}
  {Revisiting pre-trained models for {C}hinese natural language processing}.
\newblock In \emph{Proceedings of the 2020 Conference on Empirical Methods in
  Natural Language Processing: Findings}, pages 657--668, Online. Association
  for Computational Linguistics.

\bibitem[{Devlin et~al.(2019)Devlin, Chang, Lee, and
  Toutanova}]{devlin2018bert}
Jacob Devlin, Ming-Wei Chang, Kenton Lee, and Kristina Toutanova. 2019.
\newblock Bert: Pre-training of deep bidirectional transformers for language
  understanding.
\newblock In \emph{Proceedings of the 2019 Conference of the North American
  Chapter of the Association for Computational Linguistics: Human Language
  Technologies, Volume 1 (Long and Short Papers)}, pages 4171--4186.

\bibitem[{Emelin et~al.(2020)Emelin, Bras, Hwang, Forbes, and
  Choi}]{emelin2020moral}
Denis Emelin, Ronan~Le Bras, Jena~D Hwang, Maxwell Forbes, and Yejin Choi.
  2020.
\newblock Moral stories: Situated reasoning about norms, intents, actions, and
  their consequences.
\newblock \emph{arXiv preprint arXiv:2012.15738}.

\bibitem[{Fan et~al.(2018)Fan, Lewis, and Dauphin}]{fan2018hierarchical}
Angela Fan, Mike Lewis, and Yann Dauphin. 2018.
\newblock Hierarchical neural story generation.
\newblock In \emph{Proceedings of the 56th Annual Meeting of the Association
  for Computational Linguistics (Volume 1: Long Papers)}, pages 889--898.

\bibitem[{Finlayson(2012)}]{finlayson2012learning}
Mark Mark~Alan Finlayson. 2012.
\newblock \emph{Learning narrative structure from annotated folktales}.
\newblock Ph.D. thesis, Massachusetts Institute of Technology.

\bibitem[{Fleiss and Joseph(1971)}]{Fleiss1971Measuring}
Fleiss and L.~Joseph. 1971.
\newblock Measuring nominal scale agreement among many raters.
\newblock \emph{Psychological Bulletin}, 76(5):378--382.

\bibitem[{Forbes et~al.(2020)Forbes, Hwang, Shwartz, Sap, and
  Choi}]{forbes-etal-2020-social}
Maxwell Forbes, Jena~D. Hwang, Vered Shwartz, Maarten Sap, and Yejin Choi.
  2020.
\newblock \href {https://doi.org/10.18653/v1/2020.emnlp-main.48} {Social
  chemistry 101: Learning to reason about social and moral norms}.
\newblock In \emph{Proceedings of the 2020 Conference on Empirical Methods in
  Natural Language Processing (EMNLP)}, pages 653--670, Online. Association for
  Computational Linguistics.

\bibitem[{Fulgoni et~al.(2016)Fulgoni, Carpenter, Ungar, and
  Preo{\c{t}}iuc-Pietro}]{fulgoni2016empirical}
Dean Fulgoni, Jordan Carpenter, Lyle Ungar, and Daniel Preo{\c{t}}iuc-Pietro.
  2016.
\newblock An empirical exploration of moral foundations theory in partisan news
  sources.
\newblock In \emph{Proceedings of the Tenth International Conference on
  Language Resources and Evaluation (LREC'16)}, pages 3730--3736.

\bibitem[{Gehring et~al.(2017)Gehring, Auli, Grangier, Yarats, and
  Dauphin}]{gehring2017convolutional}
Jonas Gehring, Michael Auli, David Grangier, Denis Yarats, and Yann~N Dauphin.
  2017.
\newblock Convolutional sequence to sequence learning.
\newblock In \emph{International Conference on Machine Learning}, pages
  1243--1252. PMLR.

\bibitem[{Goodfellow et~al.(2016)Goodfellow, Bengio, and
  Courville}]{goodfellow2016deep}
Ian Goodfellow, Yoshua Bengio, and Aaron Courville. 2016.
\newblock \emph{Deep learning}.
\newblock MIT press.

\bibitem[{Guan et~al.(2022)Guan, Feng, Chen, He, Mao, Fan, and
  Huang}]{guan2021lot}
Jian Guan, Zhuoer Feng, Yamei Chen, Ruilin He, Xiaoxi Mao, Changjie Fan, and
  Minlie Huang. 2022.
\newblock \href {https://doi.org/10.1162/tacl_a_00469} {{LOT: A Story-Centric
  Benchmark for Evaluating Chinese Long Text Understanding and Generation}}.
\newblock \emph{Transactions of the Association for Computational Linguistics},
  10:434--451.

\bibitem[{Guan et~al.(2020)Guan, Huang, Zhao, Zhu, and
  Huang}]{guan2020knowledge}
Jian Guan, Fei Huang, Zhihao Zhao, Xiaoyan Zhu, and Minlie Huang. 2020.
\newblock A knowledge-enhanced pretraining model for commonsense story
  generation.
\newblock \emph{Transactions of the Association for Computational Linguistics},
  8:93--108.

\bibitem[{Guan and Huang(2020)}]{guan2020union}
Jian Guan and Minlie Huang. 2020.
\newblock \href {https://doi.org/10.18653/v1/2020.emnlp-main.736} {{UNION:} an
  unreferenced metric for evaluating open-ended story generation}.
\newblock In \emph{Proceedings of the 2020 Conference on Empirical Methods in
  Natural Language Processing, {EMNLP} 2020, Online, November 16-20, 2020},
  pages 9157--9166. Association for Computational Linguistics.

\bibitem[{Guan et~al.(2019)Guan, Wang, and Huang}]{guan2019story}
Jian Guan, Yansen Wang, and Minlie Huang. 2019.
\newblock Story ending generation with incremental encoding and commonsense
  knowledge.
\newblock In \emph{Proceedings of the AAAI Conference on Artificial
  Intelligence}, volume~33, pages 6473--6480.

\bibitem[{Gururangan et~al.(2020)Gururangan, Marasovi{\'c}, Swayamdipta, Lo,
  Beltagy, Downey, and Smith}]{gururangan2020don}
Suchin Gururangan, Ana Marasovi{\'c}, Swabha Swayamdipta, Kyle Lo, Iz~Beltagy,
  Doug Downey, and Noah~A Smith. 2020.
\newblock Don't stop pretraining: Adapt language models to domains and tasks.
\newblock In \emph{Proceedings of the 58th Annual Meeting of the Association
  for Computational Linguistics}, pages 8342--8360.

\bibitem[{Haidt and Joseph(2004)}]{haidt2004intuitive}
Jonathan Haidt and Craig Joseph. 2004.
\newblock Intuitive ethics: How innately prepared intuitions generate
  culturally variable virtues.
\newblock \emph{Daedalus}, 133(4):55--66.

\bibitem[{Hendrycks et~al.(2021)Hendrycks, Burns, Basart, Critch, Li, Song, and
  Steinhardt}]{hendrycks2021aligning}
Dan Hendrycks, Collin Burns, Steven Basart, Andrew Critch, Jerry Li, Dawn Song,
  and Jacob Steinhardt. 2021.
\newblock \href {https://openreview.net/forum?id=dNy_RKzJacY} {Aligning
  {\{}ai{\}} with shared human values}.
\newblock In \emph{International Conference on Learning Representations}.

\bibitem[{Hoover et~al.(2020)Hoover, Portillo-Wightman, Yeh, Havaldar, Davani,
  Lin, Kennedy, Atari, Kamel, Mendlen, Moreno, Park, Chang, Chin, Leong, Leung,
  Mirinjian, and Dehghani}]{doi:10.1177/1948550619876629}
Joe Hoover, Gwenyth Portillo-Wightman, Leigh Yeh, Shreya Havaldar,
  Aida~Mostafazadeh Davani, Ying Lin, Brendan Kennedy, Mohammad Atari, Zahra
  Kamel, Madelyn Mendlen, Gabriela Moreno, Christina Park, Tingyee~E. Chang,
  Jenna Chin, Christian Leong, Jun~Yen Leung, Arineh Mirinjian, and Morteza
  Dehghani. 2020.
\newblock \href {https://doi.org/10.1177/1948550619876629} {Moral foundations
  twitter corpus: A collection of 35k tweets annotated for moral sentiment}.
\newblock \emph{Social Psychological and Personality Science},
  11(8):1057--1071.

\bibitem[{Ianinska and Garcia-Zamor(2006)}]{ianinska2006morals}
Silvana Ianinska and Jean-Claude Garcia-Zamor. 2006.
\newblock Morals, ethics, and integrity: How codes of conduct contribute to
  ethical adult education practice.
\newblock \emph{Public Organization Review}, 6(1):3--20.

\bibitem[{Jiang et~al.(2021)Jiang, Hwang, Bhagavatula, Bras, Forbes, Borchardt,
  Liang, Etzioni, Sap, and Choi}]{jiang2021delphi}
Liwei Jiang, Jena~D Hwang, Chandra Bhagavatula, Ronan~Le Bras, Maxwell Forbes,
  Jon Borchardt, Jenny Liang, Oren Etzioni, Maarten Sap, and Yejin Choi. 2021.
\newblock Delphi: Towards machine ethics and norms.
\newblock \emph{arXiv preprint arXiv:2110.07574}.

\bibitem[{Johnson and Goldwasser(2018)}]{johnson2018classification}
Kristen Johnson and Dan Goldwasser. 2018.
\newblock Classification of moral foundations in microblog political discourse.
\newblock In \emph{Proceedings of the 56th Annual Meeting of the Association
  for Computational Linguistics (Volume 1: Long Papers)}, pages 720--730.

\bibitem[{Kong et~al.(2021)Kong, Huang, Tung, Guan, and
  Huang}]{kong-etal-2021-stylized}
Xiangzhe Kong, Jialiang Huang, Ziquan Tung, Jian Guan, and Minlie Huang. 2021.
\newblock \href {https://doi.org/10.18653/v1/2021.findings-acl.215} {Stylized
  story generation with style-guided planning}.
\newblock In \emph{Findings of the Association for Computational Linguistics:
  ACL-IJCNLP 2021}, pages 2430--2436, Online. Association for Computational
  Linguistics.

\bibitem[{Lewis et~al.(2020)Lewis, Liu, Goyal, Ghazvininejad, Mohamed, Levy,
  Stoyanov, and Zettlemoyer}]{bart}
Mike Lewis, Yinhan Liu, Naman Goyal, Marjan Ghazvininejad, Abdelrahman Mohamed,
  Omer Levy, Veselin Stoyanov, and Luke Zettlemoyer. 2020.
\newblock \href {https://www.aclweb.org/anthology/2020.acl-main.703/} {{BART:}
  denoising sequence-to-sequence pre-training for natural language generation,
  translation, and comprehension}.
\newblock In \emph{Proceedings of the 58th Annual Meeting of the Association
  for Computational Linguistics, {ACL} 2020, Online, July 5-10, 2020}, pages
  7871--7880. Association for Computational Linguistics.

\bibitem[{Li et~al.(2016)Li, Galley, Brockett, Gao, and
  Dolan}]{li2016diversity}
Jiwei Li, Michel Galley, Chris Brockett, Jianfeng Gao, and William~B Dolan.
  2016.
\newblock A diversity-promoting objective function for neural conversation
  models.
\newblock In \emph{Proceedings of the 2016 Conference of the North American
  Chapter of the Association for Computational Linguistics: Human Language
  Technologies}, pages 110--119.

\bibitem[{Lin(2004)}]{lin-2004-rouge}
Chin-Yew Lin. 2004.
\newblock \href {https://www.aclweb.org/anthology/W04-1013} {{ROUGE}: A package
  for automatic evaluation of summaries}.
\newblock In \emph{Text Summarization Branches Out}, pages 74--81, Barcelona,
  Spain. Association for Computational Linguistics.

\bibitem[{Lin et~al.(2021)Lin, Meng, Sun, Han, Kuang, Li, and
  Wu}]{lin2021bertgcn}
Yuxiao Lin, Yuxian Meng, Xiaofei Sun, Qinghong Han, Kun Kuang, Jiwei Li, and
  Fei Wu. 2021.
\newblock Bertgcn: Transductive text classification by combining gcn and bert.
\newblock \emph{arXiv preprint arXiv:2105.05727}.

\bibitem[{Liu et~al.(2019)Liu, Ott, Goyal, Du, Joshi, Chen, Levy, Lewis,
  Zettlemoyer, and Stoyanov}]{liu2019roberta}
Yinhan Liu, Myle Ott, Naman Goyal, Jingfei Du, Mandar Joshi, Danqi Chen, Omer
  Levy, Mike Lewis, Luke Zettlemoyer, and Veselin Stoyanov. 2019.
\newblock Roberta: A robustly optimized bert pretraining approach.
\newblock \emph{arXiv preprint arXiv:1907.11692}.

\bibitem[{Loper and Bird(2002)}]{loper2002nltk}
Edward Loper and Steven Bird. 2002.
\newblock Nltk: The natural language toolkit.
\newblock In \emph{Proceedings of the ACL-02 Workshop on Effective Tools and
  Methodologies for Teaching Natural Language Processing and Computational
  Linguistics}, pages 63--70.

\bibitem[{Louis and Sutton(2018)}]{louis2018deep}
Annie Louis and Charles Sutton. 2018.
\newblock Deep dungeons and dragons: Learning character-action interactions
  from role-playing game transcripts.
\newblock In \emph{Proceedings of the 2018 Conference of the North American
  Chapter of the Association for Computational Linguistics: Human Language
  Technologies, Volume 2 (Short Papers)}, pages 708--713.

\bibitem[{Lourie et~al.(2021)Lourie, Le~Bras, and Choi}]{lourie2021scruples}
Nicholas Lourie, Ronan Le~Bras, and Yejin Choi. 2021.
\newblock Scruples: A corpus of community ethical judgments on 32, 000
  real-life anecdotes.
\newblock In \emph{Proceedings of the AAAI Conference on Artificial
  Intelligence}, volume~35, pages 13470--13479.

\bibitem[{Mostafazadeh et~al.(2016)Mostafazadeh, Chambers, He, Parikh, Batra,
  Vanderwende, Kohli, and Allen}]{mostafazadeh2016corpus}
Nasrin Mostafazadeh, Nathanael Chambers, Xiaodong He, Devi Parikh, Dhruv Batra,
  Lucy Vanderwende, Pushmeet Kohli, and James Allen. 2016.
\newblock A corpus and cloze evaluation for deeper understanding of commonsense
  stories.
\newblock In \emph{Proceedings of NAACL-HLT}, pages 839--849.

\bibitem[{Narayan et~al.(2018)Narayan, Cohen, and Lapata}]{narayan2018don}
Shashi Narayan, Shay~B Cohen, and Mirella Lapata. 2018.
\newblock Don't give me the details, just the summary! topic-aware
  convolutional neural networks for extreme summarization.
\newblock In \emph{Proceedings of the 2018 Conference on Empirical Methods in
  Natural Language Processing}, pages 1797--1807.

\bibitem[{Papineni et~al.(2002)Papineni, Roukos, Ward, and
  Zhu}]{papineni2002bleu}
Kishore Papineni, Salim Roukos, Todd Ward, and Wei-Jing Zhu. 2002.
\newblock Bleu: a method for automatic evaluation of machine translation.
\newblock In \emph{Proceedings of the 40th annual meeting of the Association
  for Computational Linguistics}, pages 311--318.

\bibitem[{Pavan et~al.(2020)Pavan, Dos~Santos, Lan, Martins, Santos, Deutsch,
  Costa, Hsieh, and Paraboni}]{pavan2020morality}
Matheus~C Pavan, Vitor~G Dos~Santos, Alex~GJ Lan, Joao Martins, Wesley~R
  Santos, Caio Deutsch, Pablo~B Costa, Fernando~C Hsieh, and Ivandre Paraboni.
  2020.
\newblock Morality classification in natural language text.
\newblock \emph{IEEE Transactions on Affective Computing}.

\bibitem[{Radford et~al.(2019)Radford, Wu, Child, Luan, Amodei, and
  Sutskever}]{radford2019language}
Alec Radford, Jeffrey Wu, Rewon Child, David Luan, Dario Amodei, and Ilya
  Sutskever. 2019.
\newblock Language models are unsupervised multitask learners.
\newblock \emph{OpenAI blog}, 1(8):9.

\bibitem[{Rae et~al.(2020)Rae, Potapenko, Jayakumar, Hillier, and
  Lillicrap}]{Rae2020Compressive}
Jack~W. Rae, Anna Potapenko, Siddhant~M. Jayakumar, Chloe Hillier, and
  Timothy~P. Lillicrap. 2020.
\newblock \href {https://openreview.net/forum?id=SylKikSYDH} {Compressive
  transformers for long-range sequence modelling}.
\newblock In \emph{International Conference on Learning Representations}.

\bibitem[{Raffel et~al.(2020)Raffel, Shazeer, Roberts, Lee, Narang, Matena,
  Zhou, Li, and Liu}]{raffel2020exploring}
Colin Raffel, Noam Shazeer, Adam Roberts, Katherine Lee, Sharan Narang, Michael
  Matena, Yanqi Zhou, Wei Li, and Peter~J Liu. 2020.
\newblock Exploring the limits of transfer learning with a unified text-to-text
  transformer.
\newblock \emph{Journal of Machine Learning Research}, 21:1--67.

\bibitem[{Rashkin et~al.(2020)Rashkin, Celikyilmaz, Choi, and
  Gao}]{rashkin2020plotmachines}
Hannah Rashkin, Asli Celikyilmaz, Yejin Choi, and Jianfeng Gao. 2020.
\newblock Plotmachines: Outline-conditioned generation with dynamic plot state
  tracking.
\newblock In \emph{Proceedings of the 2020 Conference on Empirical Methods in
  Natural Language Processing (EMNLP)}, pages 4274--4295.

\bibitem[{Ribeiro et~al.(2020)Ribeiro, Wu, Guestrin, and
  Singh}]{ribeiro-etal-2020-beyond}
Marco~Tulio Ribeiro, Tongshuang Wu, Carlos Guestrin, and Sameer Singh. 2020.
\newblock \href {https://doi.org/10.18653/v1/2020.acl-main.442} {Beyond
  accuracy: Behavioral testing of {NLP} models with {C}heck{L}ist}.
\newblock In \emph{Proceedings of the 58th Annual Meeting of the Association
  for Computational Linguistics}, pages 4902--4912, Online. Association for
  Computational Linguistics.

\bibitem[{Rose et~al.(2010)Rose, Engel, Cramer, and Cowley}]{rose2010automatic}
Stuart Rose, Dave Engel, Nick Cramer, and Wendy Cowley. 2010.
\newblock Automatic keyword extraction from individual documents.
\newblock \emph{Text mining: applications and theory}, 1:1--20.

\bibitem[{Sap et~al.(2020)Sap, Gabriel, Qin, Jurafsky, Smith, and
  Choi}]{sap-etal-2020-social}
Maarten Sap, Saadia Gabriel, Lianhui Qin, Dan Jurafsky, Noah~A. Smith, and
  Yejin Choi. 2020.
\newblock \href {https://doi.org/10.18653/v1/2020.acl-main.486} {Social bias
  frames: Reasoning about social and power implications of language}.
\newblock In \emph{Proceedings of the 58th Annual Meeting of the Association
  for Computational Linguistics}, pages 5477--5490, Online. Association for
  Computational Linguistics.

\bibitem[{Shao et~al.(2019)Shao, Huang, Wen, Xu, and Zhu}]{shao-etal-2019-long}
Zhihong Shao, Minlie Huang, Jiangtao Wen, Wenfei Xu, and Xiaoyan Zhu. 2019.
\newblock \href {https://doi.org/10.18653/v1/D19-1321} {Long and diverse text
  generation with planning-based hierarchical variational model}.
\newblock In \emph{Proceedings of the 2019 Conference on Empirical Methods in
  Natural Language Processing and the 9th International Joint Conference on
  Natural Language Processing (EMNLP-IJCNLP)}, pages 3257--3268, Hong Kong,
  China. Association for Computational Linguistics.

\bibitem[{Stab and Gurevych(2017)}]{stab2017parsing}
Christian Stab and Iryna Gurevych. 2017.
\newblock Parsing argumentation structures in persuasive essays.
\newblock \emph{Computational Linguistics}, 43(3):619--659.

\bibitem[{Tappan and Brown(1989)}]{tappan1989stories}
Mark Tappan and Lyn~Mikel Brown. 1989.
\newblock Stories told and lessons learned: Toward a narrative approach to
  moral development and moral education.
\newblock \emph{Harvard Educational Review}, 59(2):182--206.

\bibitem[{Vitz(1990)}]{vitz1990use}
Paul~C Vitz. 1990.
\newblock The use of stories in moral development: New psychological reasons
  for an old education method.
\newblock \emph{American psychologist}, 45(6):709.

\bibitem[{Volkova et~al.(2017)Volkova, Shaffer, Jang, and
  Hodas}]{volkova2017separating}
Svitlana Volkova, Kyle Shaffer, Jin~Yea Jang, and Nathan Hodas. 2017.
\newblock Separating facts from fiction: Linguistic models to classify
  suspicious and trusted news posts on twitter.
\newblock In \emph{Proceedings of the 55th Annual Meeting of the Association
  for Computational Linguistics (Volume 2: Short Papers)}, pages 647--653.

\bibitem[{Wang and Wan(2019)}]{DBLP:conf/ijcai/Wang019b}
Tianming Wang and Xiaojun Wan. 2019.
\newblock \href {https://doi.org/10.24963/ijcai.2019/727} {{T-CVAE:}
  transformer-based conditioned variational autoencoder for story completion}.
\newblock In \emph{Proceedings of the Twenty-Eighth International Joint
  Conference on Artificial Intelligence, {IJCAI} 2019, Macao, China, August
  10-16, 2019}, pages 5233--5239. ijcai.org.

\bibitem[{Wolf et~al.(2020)Wolf, Chaumond, Debut, Sanh, Delangue, Moi, Cistac,
  Funtowicz, Davison, Shleifer et~al.}]{wolf2020transformers}
Thomas Wolf, Julien Chaumond, Lysandre Debut, Victor Sanh, Clement Delangue,
  Anthony Moi, Pierric Cistac, Morgan Funtowicz, Joe Davison, Sam Shleifer,
  et~al. 2020.
\newblock Transformers: State-of-the-art natural language processing.
\newblock In \emph{Proceedings of the 2020 Conference on Empirical Methods in
  Natural Language Processing: System Demonstrations}, pages 38--45.

\bibitem[{Yao et~al.(2019)Yao, Peng, Weischedel, Knight, Zhao, and
  Yan}]{yao2019plan}
Lili Yao, Nanyun Peng, Ralph Weischedel, Kevin Knight, Dongyan Zhao, and Rui
  Yan. 2019.
\newblock Plan-and-write: Towards better automatic storytelling.
\newblock In \emph{Proceedings of the AAAI Conference on Artificial
  Intelligence}, volume~33, pages 7378--7385.

\bibitem[{Zhang et~al.(2019)Zhang, Kishore, Wu, Weinberger, and
  Artzi}]{zhang2019bertscore}
Tianyi Zhang, Varsha Kishore, Felix Wu, Kilian~Q Weinberger, and Yoav Artzi.
  2019.
\newblock Bertscore: Evaluating text generation with bert.
\newblock In \emph{International Conference on Learning Representations}.

\bibitem[{Zhao et~al.(2017)Zhao, Zhao, and Eskenazi}]{zhao2017learning}
Tiancheng Zhao, Ran Zhao, and Maxine Eskenazi. 2017.
\newblock Learning discourse-level diversity for neural dialog models using
  conditional variational autoencoders.
\newblock In \emph{Proceedings of the 55th Annual Meeting of the Association
  for Computational Linguistics (Volume 1: Long Papers)}, pages 654--664.

\bibitem[{Zhao et~al.(2019)Zhao, Chen, Zhang, Zhao, Liu, Lu, Chen, Deng, Ju,
  and Du}]{zhao2019uer}
Zhe Zhao, Hui Chen, Jinbin Zhang, Xin Zhao, Tao Liu, Wei Lu, Xi~Chen, Haotang
  Deng, Qi~Ju, and Xiaoyong Du. 2019.
\newblock Uer: An open-source toolkit for pre-training models.
\newblock \emph{EMNLP-IJCNLP 2019}, page 241.

\end{thebibliography}
\bibliographystyle{acl_natbib}

\appendix

\section{\textsc{storal} Construction}
\subsection{Data Source}\label{source}
We show the full list of web pages used for constructing \textsc{storal} in Table~\ref{tab:data_source}. We initially collect 52,017 Chinese and 2,630 English raw texts from the web pages. Then we de-duplicate the texts by removing those texts which overlap with others more than twenty words. After de-duplication, we finally collected 19,197 Chinese and 2,598 English texts. And we construct \textsc{storal} based on these texts.
\subsection{Data Annotation}\label{constraints}
Table~\ref{tab:st}/\ref{tab:mo} shows the examples given to the annotators to inform them of the requirements for stories/morals, respectively. If the constraints are not met, we ask annotators to refine the story and moral.  All workers were paid more than \$7 per hour on average. %We payed \$0.3 on average for annotating each example in \textsc{storal}. We decided the payment according to the average text length. 
%We show an annotation example in Table~\ref{tab:ant_example}.

\begin{table}[!ht]
\small
    \centering
    \begin{tabular}{p{207pt}}
    \toprule
    \textbf{Example 1:} \textit{Come on Bear! What a beautiful day! Go for a walk with your father! Take a deep breath and smell the flowers. But don't pick the flowers. Listen to the birds sing. But don't scare them. %Look, your uncle is working. Give him a wave. %They are selling balloons over there! Pick one! 
    How beautiful the world is. Isn't it, dear Bear?}\\ %\textit{When you are boring, take the subway during the busy morning rush hour. You will be relieved at once. And meanwhile you may be more devastated. $\cdots$ No matter how much passion and energy you build up, it will wear away day after day.} \\%or whether it is really eaten by termites. In this season of rippling green life, it announces its own death in a bipey form.\\
    \midrule
    \textbf{Example 2:} \textit{When I was a child, I heard a story that felt very regrettable. I felt sorry for the protagonist of the story.} Long ago, there lived %a beautiful little girl in the West Village. %Because she was so beautiful, people forgot her name and called her ``Beauty''. 
    $\cdots$ Such trees are now found all over Uganda. \\%Some call it the flame tree, while some call it the lover tree. \\%It's a beautiful and moving story, but it's also a story of believing in what you believe.\\
    \midrule
    \textbf{Example 3:} I have a well-off friend. When she first entered college, she had many good wishes and thought she could achieve her goals. $\cdots$ Now she felt very painful under the strong mental pressure. \textit{I can understand her feelings. $\cdots$ If magnifying your own pain, you will get trapped in the mire of your pain, and even feel that life is too unfair to you.}\\% Once people are disillusioned, it is easy to go to extremes. If you shrink your pain infinitely, you can even see it as a step on the road. When you see it, climb it, and it becomes the stepping stone to new heights. I remember a writer once said: either you drive life, or life drives you, your attitude determines who is the mount and who is the rider.}
    \midrule
    \textbf{Example 4:} \textit{Raul sat at his door, frowning. $\cdots$ His father told Raul a true story:} A wild wolf escaped into a cave after being wounded by a hunter's arrow. $\cdots$ \textit{After hearing the story, Raul cheered up immediately. $\cdots$}\\
   \bottomrule
    \end{tabular}
    \caption{Examples of stories provided for the annotators. Each example does not meet one of the following requirements in order: (1) having a clear beginning and ending; (2) not stating anything irrelevant to the main plot; (3) not stating any explicit arguments for the moral; and (4) not telling the story in a nested form. The sentences causing the above issues are in \textit{italic}.}
    \label{tab:st}
\end{table}

\begin{table}[!ht]
\small
    \centering
    \begin{tabular}{p{207pt}}
    \toprule
    \textbf{Example 1:} \textit{If you saw a thief in a crowded bus, would you bravely stop him?} Please reflect on yourself instead of just complaining that our world is becoming worse. \textit{Without the foothold for dirt, the flower of civilization is bound to be fragrant.}\\
    \midrule
    \textbf{Example 2:} \textit{The story tells us: we should remember that} we should become a polite person and communicate with others carefully.\\ 
    \midrule
    \textbf{Example 3:} As long as you keep your sanity and make right judgments, all the barriers will not become an obstacle, \textit{just like the beautiful girl in the story}.\\

   \bottomrule
    \end{tabular}
    \caption{Examples of morals provided for the annotators. Each of the examples does not satisfy one of the following constraints in order: (1) describing only the main standpoint and not stating any sub-arguments or proofs, and (2) not stating anything irrelevant to the moral, and (3) not involving any specific characters in the story. We highlight the sentences leading to the above issues in \textit{italic}.} 
    \label{tab:mo}
\end{table}

\begin{table*}[!th]
\tiny
    \centering
    \begin{tabular}{lr}
    \toprule
    %\multicolumn{2}{c}{\textbf{\textsc{storal-zh}}}\\
    %\midrule
    \textbf{Links}&\textbf{Number}\\
    \midrule
    \url{http://www.qbaobei.com/jiaoyu/yegs/yygs/}&14,674\\
    \url{https://www.517gj.com/yuyangushi/}&14,474\\
    \url{https://www.etgushi.com/zgyy/}&6,691\\
    \url{https://www.chazidian.com/gushi_1/}&3,457\\
    \url{http://www.feel-bar.com}&3,329\\
    \url{http://www.xiaole8.com/renshengzheli/}&2,509\\
    \url{http://www.zuowen.com/sucai/zheli/}&2,421\\
    \url{http://www.rensheng5.com}&2,092\\
    \url{https://www.yuyangushi.com/lz/xgsddl}&1,886\\
    \url{http://www.gushi88.cn/ErTong/ZhongGuoYuYan_1}&484\\
    \midrule
    Grand Total&52,017\\
    \midrule
    %\multicolumn{2}{c}{\textbf{\textsc{storal-en}}}\\
    \midrule
    \textbf{Links}&\textbf{Number}\\
    \midrule
    \url{https://moralstories26.com}&799\\
    \url{https://english.7139.com/2539/}&552\\
    %\url{https://freestoriesforkids.com/short-stories}&287\\
    \url{https://kidsfables.com}&193\\
    %\url{http://m.enread.com/index.php?mid=2&catid=27}&148\\
    \url{http://read.gov/aesop}&145\\
    %\url{https://www.moralstories.org}&129\\
    \url{http://www.taleswithmorals.com}&108\\
    \url{https://www.studentuk.com/category/fable}&101\\
    \url{http://www.english-for-students.com/Moral-Stories.html}&97\\
    %\url{https://www.kidsgen.com/moral_stories}&83\\
    \url{https://www.advance-africa.com/English-Moral-Stories.html}&65\\
    \url{https://www.gutenberg.org/files/25512/25512-h/25512-h.htm}&52\\
    {Others}&518\\
    \midrule
    Grand Total&2,630\\
    \bottomrule    
    \end{tabular}
    \caption{List of source web pages used for constructing \textsc{storal-zh}~(Top) and \textsc{storal-en}~(Bottom). Numbers in the right column means the number of raw texts initially collected from the corresponding web page.} %\textit{Others} means that these web sites have been shut down.}
    \label{tab:data_source}
\end{table*}

\begin{figure*}[!th]
\centering
\subfigure[Stories in \textsc{storal-zh}]{\includegraphics[width=0.49\linewidth]{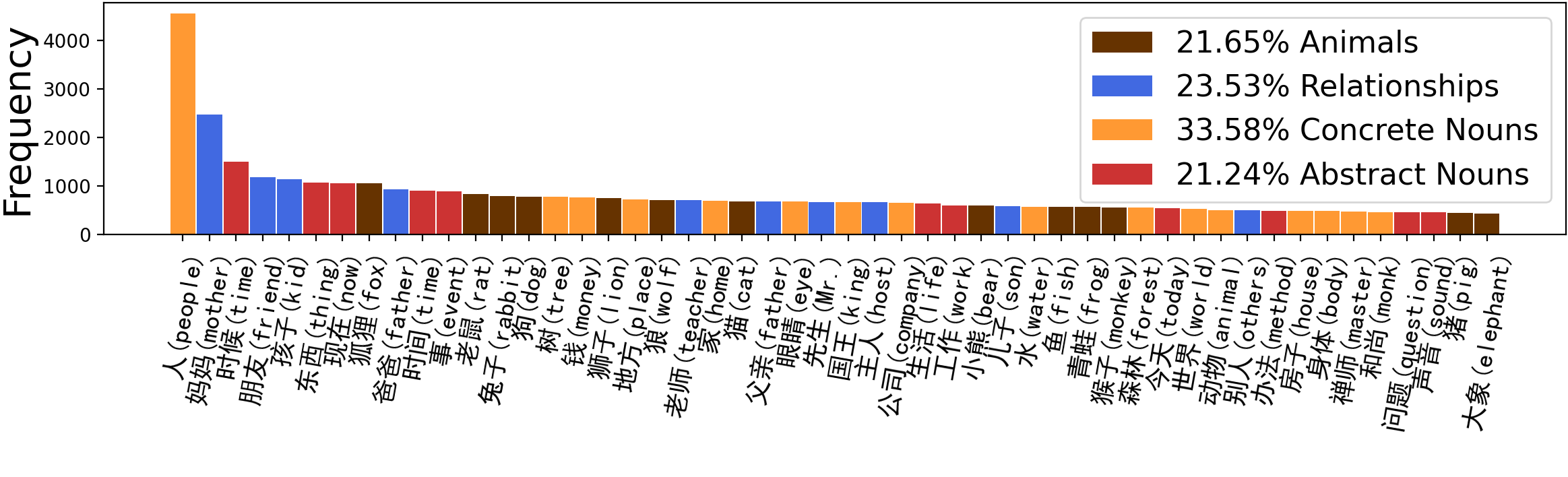}}
\subfigure[Morals in \textsc{storal-zh}]{\includegraphics[width=0.47\linewidth]{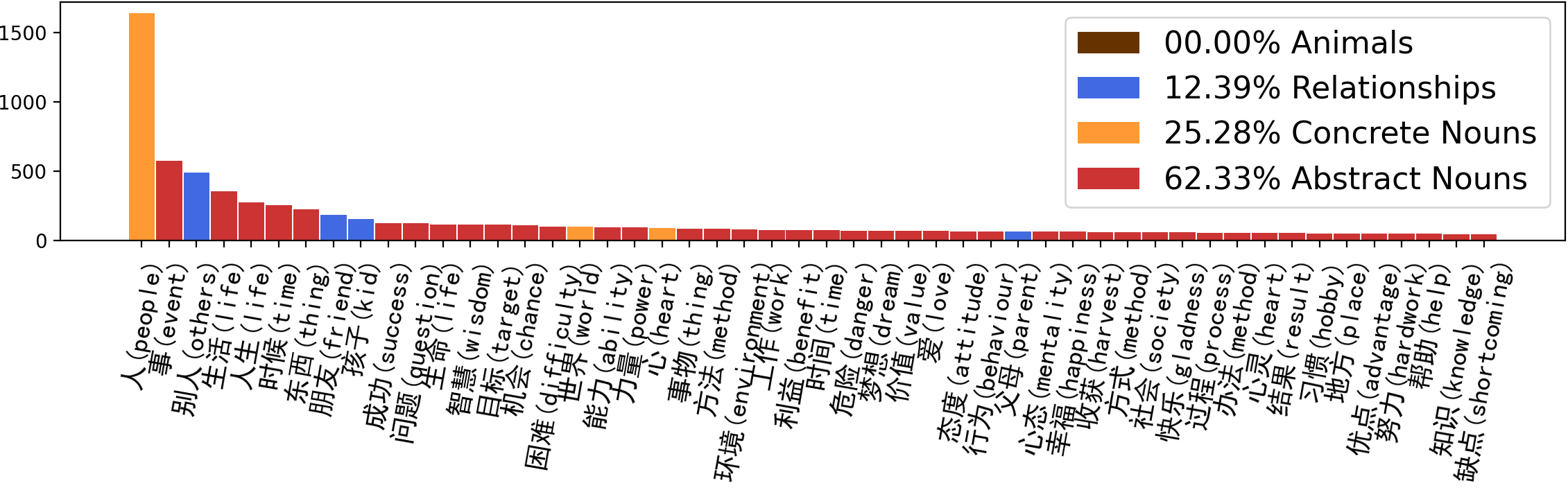}}
\subfigure[Stories in \textsc{storal-en}]{\includegraphics[width=0.49\linewidth]{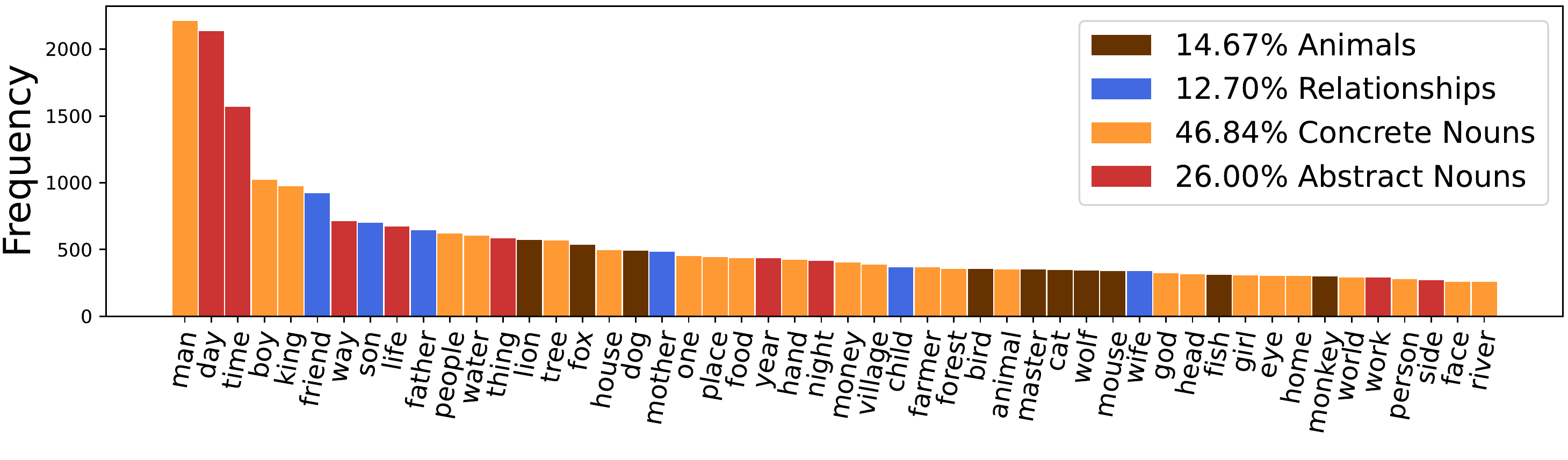}}
\subfigure[Morals in \textsc{storal-en}]{\includegraphics[width=0.47\linewidth]{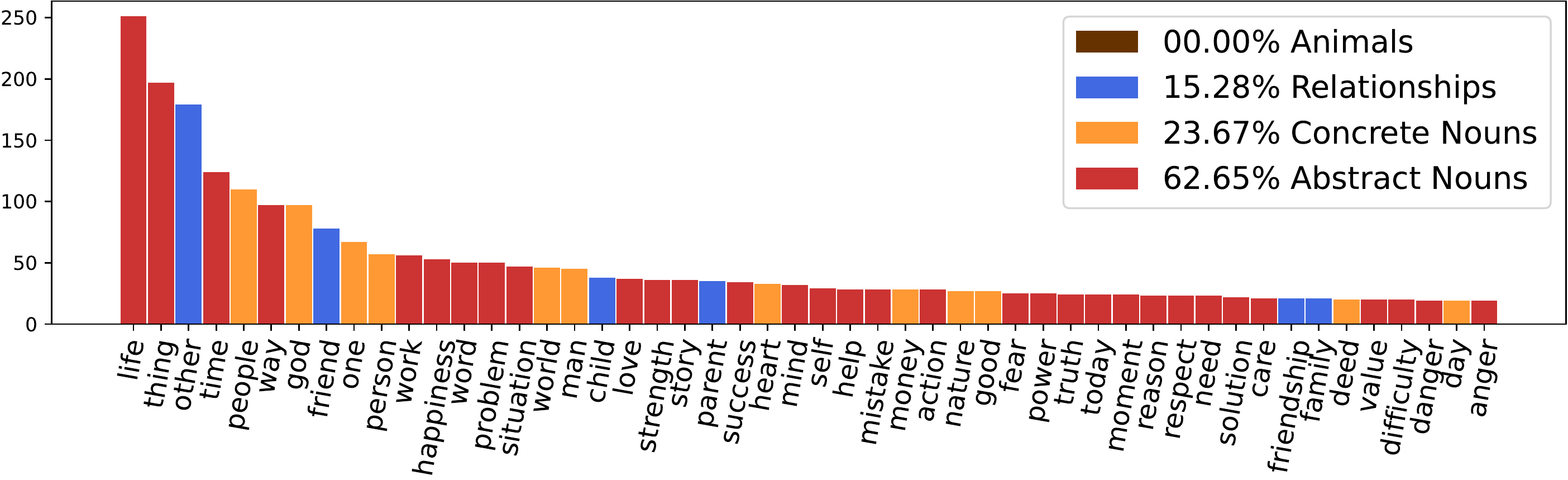}}
%\subfigure[$\pi_{\text{max}}(10,000)$]{\includegraphics[width=2.73in]{figures/max.pdf}}
%\subfigure[$\pi_{\text{mix}}(10,000)$]{\includegraphics[width=2.73in]{figures/mix.pdf}}
\caption{Top 50 most frequent nouns for stories and morals in \textsc{storal-zh} and \textsc{storal-en}. 
The numbers in the legend show the percentages of the total frequency of the nouns of the same type among the 50 nouns.} %We do not show the situation for {\textsc{storal-zh}} since it exhibits a similar distribution of the noun types.}
\label{topic}
\end{figure*}

\subsection{Analysis of High-Frequency Words}\label{high-freq} %To investigate the level of abstraction of morals in \textsc{storal}, 
To investigate the topic features of \textsc{storal},
we count the top 50 most frequent nouns in \textsc{storal}~(excluding stop words) as shown in Figure~\ref{topic}. We roughly categorize these words into four types: \textbf{(1) Animals}: animals are popular as protagonists in moral stories since they usually have various but clear characteristics~(e.g., ``sly foxes''), which embody rich commonsense knowledge; \textbf{(2) Relationships}: such nouns are used to describe the inter-character relationships in a story~(e.g., {``friend''}), which are useful for modeling characters' motivation and behavior; \textbf{(3) Concrete nouns}: they refer to physical entities that can be observed, such as {{``water''}}; and \textbf{\textbf{(4) Abstract nouns}}: they refer to abstract concepts, such as {``difficulty''}. We manually check the proportional distribution of the four types for stories and morals, respectively. The results in Figure~\ref{topic} demonstrate that morals contain significantly less concrete nouns and more abstract nouns than stories. And morals contain little animal words but almost as many relationship words as stories, indicating that morals may be independent of specific characters but relate to general interpersonal relations. The result shows that morals are more abstract than stories.%The most frequent noun {``man''}/{``life''} in stories/morals of \textsc{storal-en} comprises 2.3\%/4.3\% of all nouns~(the proportions are 1.6\%/7.8\% for \textsc{storal-zh}), 
%which illustrates the rich diversity of \textsc{storal}.

\begin{table}[!t]
\small
    \centering
    \begin{tabular}{ll}
    \toprule
    \textbf{Stories}&\textbf{Morals}\\
    %\multicolumn{l}{c}{\textbf{Stories}}& \multicolumn{l}{c}{\textbf{Morals}}\\
    \midrule
    \multicolumn{2}{l}{\textbf{Dataset: \textsc{storal-zh}}}\\
    \midrule
%\begin{CJK}{UTF8}{gbsn}\scriptsize走 着 走 着\end{CJK}\newline (as one is walking)&\begin{CJK}{UTF8}{gbsn}\scriptsize我们 要 做 一个\end{CJK}\newline (we should be a)\\
as one is walking&we should be a\\
%\begin{CJK}{UTF8}{gbsn}\scriptsize对 他 说 你\end{CJK}\newline (say to him that you)&\begin{CJK}{UTF8}{gbsn}\scriptsize每个 人 都 有\end{CJK}\newline (everyone has)\\
say to him that you&everyone has\\
%\begin{CJK}{UTF8}{gbsn}\scriptsize想 了 想 说\end{CJK}\newline (say after thinking)&\begin{CJK}{UTF8}{gbsn}\scriptsize人 都 有 自己\end{CJK}\newline (everyone has his own)\\
say after thinking&everyone has his own\\
%\begin{CJK}{UTF8}{gbsn}\scriptsize是  世界 上 最\end{CJK}\newline (the most in the world)&\begin{CJK}{UTF8}{gbsn}\scriptsize都 有 自己 的\end{CJK}\newline (has own)\\
the most in the world&has own\\
%{\begin{CJK}{UTF8}{gbsn}\scriptsize小 动物 们 都\end{CJK}\newline (all the animals)}&\begin{CJK}{UTF8}{gbsn}\scriptsize我们 每个 人 都\end{CJK}\newline (each of us)\\
all the animals&each of us\\
%{\begin{CJK}{UTF8}{gbsn}\scriptsize所有 的 人 都\end{CJK}\newline (all the persons)}\\
all the persons&we should know to \\
a place far away&for anything, we should\\
the dad of the pink pig&be one who knows to \\
this is my  &for anything, be\\
in the forest there lived a &is a true\\
%{\begin{CJK}{UTF8}{gbsn}\scriptsize很 远 的 地方\end{CJK}\newline (a place far away)}&我们要懂得去\\
%{\begin{CJK}{UTF8}{gbsn}\scriptsize粉红 猪 的 爸爸\end{CJK}\newline (the dad of the pink pig)}&任何事情都要\\
%{\begin{CJK}{UTF8}{gbsn}\scriptsize这 是 我 的\end{CJK}\newline (this is my)}&要做一个懂得\\
%{\begin{CJK}{UTF8}{gbsn}\scriptsize森林里 住 着\end{CJK}\newline (in the forest there lives a)}&做任何事情都\\
\midrule
\midrule
\multicolumn{2}{l}{\textbf{Dataset: \textsc{storal-en}}}\\
\midrule
once upon a time&we should try to\\
upon a time there& the best way to\\
a time there was&it is better to\\
time there was a&it is easy to\\
there was once a &we should learn to\\
once there was a&those who help themselves\\
was not able to&with what we have\\
as soon as he&be happy with what\\
and asked him to&we should not judge\\
did n't want to&look before you leap\\
   \bottomrule
    \end{tabular}
    \caption{Top 10 most frequent 4-grams in \textsc{storal-zh} and \textsc{storal-en} respectively. The Chinese 4-grams in \textsc{stroal-zh} are translated into English.}
    \label{tab:ngram}
\end{table}

\begin{table*}[!t]
\small
    \centering
    \begin{tabular}{l|c|c}
    \toprule
    \textbf{Datasets}&\textbf{\textsc{storal-zh}}&\textbf{\textsc{storal-en}}\\
    \midrule
    \textbf{BERT}&bert-base-chinese~\cite{devlin2018bert}&bert-base-uncased~\cite{devlin2018bert}\\
    \textbf{RoBERTa}&hfl/chinese-roberta-wwm-ext~\cite{cui-etal-2020-revisiting}&roberta-base~\cite{liu2019roberta}\\
    \textbf{GPT2}&uer/gpt2-chinese-cluecorpussmall~\cite{zhao2019uer}&gpt2~\cite{radford2019language}\\
    \textbf{T5}&LongLM~\cite{guan2021lot}&t5-base~\cite{raffel2020exploring}\\
    \bottomrule
    \end{tabular}
    \caption{Names of register models used in our experiment.}
    \label{tab:model}
\end{table*}

\begin{table}[!th]
\small
    \centering
    \begin{tabular}{l|cc|cc}
    \toprule
    \multirow{2}{*}{\textbf{Models}}&\multicolumn{2}{c|}{\textbf{\textsc{st2mo}}}&\multicolumn{2}{c}{\textbf{\textsc{mo2st}}}\\
    &\textbf{\textsc{zh}}&\textbf{\textsc{en}}&\textbf{\textsc{zh}}&\textbf{\textsc{en}}\\
    \midrule
    \textbf{ConvS2S}&31.92&28.68&33.44&35.59\\
    \textbf{Fusion}&33.85&25.23&38.81&35.16\\
    \midrule
    \textbf{GPT2}&68.52&73.73&50.49&64.90\\
    \textbf{PM}&N/A&N/A&52.41&62.53\\
    \textbf{T5}&\underline{89.20}&\underline{90.57}&56.11&63.45\\
    \textbf{T5-Post}&\textbf{90.98}&\textbf{91.87}&\underline{58.58}&\textbf{75.67}\\
    \midrule
    \textbf{RA-T5}&86.50&88.69&\textbf{59.50}&\underline{74.92}\\
    \midrule
    \textbf{Truth}&\textit{77.49}&\textit{80.03}&\textit{77.49}&\textit{80.03}\\
    \bottomrule
    \end{tabular}
    \caption{Automatic moral faithfulness scores. The score of PM for the \textsc{st2mo} task is N/A since we do not experiment with PM for this task.}
    \label{tab:faith_auto}
\end{table}

Furthermore, Table~\ref{tab:ngram} shows the most frequent 4-grams in \textsc{storal}, further  indicating that morals are more abstract than stories. Each of the 4-grams in Table~\ref{tab:ngram} comprises less than $0.01\%$ of all 4-grams in the corresponding dataset, showing the diversity of \textsc{storal}.

\subsection{Discussion about \textsc{storal}}\label{discuss}
The high-quality examples in \textsc{storal} are full of commonsense and discourse relations. As exemplified in Table 1 in the main paper, the common sense is mainly regarding the characters' reaction and intention~(e.g., {``the cows dispersed''} and then the {``tiger''} and {``lion''} intend to kill them), as well as the nature of physical objects and abstract concepts~(e.g. {``cows''} may be the food of {``lions''} and {``tigers''}, and {``unity''} refers to {``keeping together for a common goal''}). Additionally, the stories usually have a specific discourse structure, %to convince readers of the morals, 
i.e., the premise to introduce the story settings~(e.g., the characters {``four cows''} and the location {``a meadow''}), the right or wrong behavior~({``stay together or not''}) and the endings~({``living well or being killed''}). We believe it is an essential topic of future work to develop a better approach to model such commonsense and discourse relations.

\section{Experiments}
\subsection{Implementation}\label{imple}
We implement the pretrained models used in our experiment mainly based on the register models of HuggingFace~\cite{wolf2020transformers}. Table~\ref{tab:model} shows the names of the used register models. Note that we use LongLM$_{\rm base}$~\cite{guan2021lot} as the T5 model for experiments on \textsc{storal-zh}, which has not been registered on HuggingFace.

All results in the main paper and the appendix are based on one NVIDIA Tesla v100 (16G memory). All reported results are based on one single running. The CPU is Intel Xeon Gold 5218. It cost less than 5 hours for fine-tuning each model on \textsc{storal}. We set the hyper-parameters following the default parameters of HuggingFace. 
%This paper states the number of algorithm runs used to compute each reported result.
%The reported results are based on once running states the number of algorithm runs used to compute each reported result.

\subsection{Automatic Evaluation for Moral Faithfulness}\label{auto_cons} 
We follow \citet{guan2020union} to train a learnable metric to evaluate moral faithfulness. Specifically, we fine-tune RoBERTa$_{\textsc{base}}$ as a classifier to distinguish whether a story matches a moral. We regard ground-truth examples as positive where the story and moral are matched, and construct negative examples by replacing the story or moral with a randomly sampled one. Finally, the classifier achieves an accuracy of 77.32/79.21\% on the data constructed based on the test set of \textsc{storal-zh}/\textsc{storal-en} respectively. Then we calculate the faithfulness score as the average classifier score of all generated texts for the inputs. 

Table~\ref{tab:faith_auto} presents the evaluation results. We can see that pretrained models achieve better faithfulness than the non-pretrained models as shown by the much higher faithfulness scores. However, we also observe that the faithfulness score of the ground-truth texts is lower than some models~(e.g., T5) when generating morals. Therefore, it is still necessary to manually evaluate faithfulness. %We summarize the reasons as follows: (a) It is still difficult for the learned classifier to judge implicit faithfulness in some ground-truth texts, which requires a strong reasoning ability that we focus on in this paper; And (b) the models tend to generate a limited set of texts~(as demonstrated by much lower distinct-4 than ground-truth texts) with generic surface forms~\cite{guan2020knowledge}~(e.g., \textit{``we should not ...''}), which may get high faithfulness scores easily. However, we believe it is still meaningful to use the faithfulness score to compare different models with similar generation diversity. 

\begin{table}[!t]
\scriptsize
    \centering
    \begin{tabular}{l|c|cc|cc}
    \toprule
    \multirow{2}{*}{\textbf{Models}}&\multirow{2}{*}{\textbf{\# P}}&\multicolumn{2}{c|}{\textbf{\textsc{MoCpt}}}&\multicolumn{2}{c}{\textbf{\textsc{MoPref}}}\\
    &&\textbf{\textsc{zh}}&\textbf{\textsc{en}}&\textbf{\textsc{zh}}&\textbf{\textsc{en}}\\
    \midrule
    %\textbf{\# Examples}&N/A&420&355&413&340\\
    %\midrule
    %\textbf{Random}&N/A&20.00&20.00&50.12&50.00\\
    \textbf{BERT w/o Story}&110M&20.71&21.69&72.64&77.62\\
    \midrule
    %\textbf{Tfmr}&40M&24.29&23.38&54.63&51.47\\
    \textbf{BERT}&110M&65.24&54.08&85.37&79.36\\
    \textbf{RoBERTa}&110M&66.90&61.69&{90.49}&80.52\\
    \textbf{RoBERTa-Post}&110M&67.14&55.77&{89.27}&{84.01}\\
    \textbf{T5}&220M&{74.52}&62.25&78.05&77.91\\
    \textbf{T5-Post}&220M&{74.05}&{67.61}&81.22&81.10\\
    \midrule
    \textbf{RA-RoBERTa}&110M&66.43&{63.94}&88.54&{86.63}\\
    \textbf{RA-T5}&220M&{74.05}&{67.61}&80.73&80.23\\
    \bottomrule
    \end{tabular}
    \caption{Accuracy~(\%) for \textsc{moCpt} and \textsc{moPref}  on the validation set.}
    \label{tab:select_acc_val}
\end{table}

\begin{table}[!t]
\scriptsize
    \centering
    \begin{tabular}{l|ccccc|ccccc}
    \toprule
\multirow{2}{*}{\textbf{Models}}&\multicolumn{2}{c}{\textbf{\textsc{st2mo}}}&\multicolumn{2}{c}{\textbf{\textsc{mo2st}}}\\
&\textbf{\textsc{zh}}&\textbf{\textsc{en}}&\textbf{\textsc{zh}}&\textbf{\textsc{en}}\\
\midrule
\textbf{Fusion}&14.44/1.80&10.78/0.92&16.06/6.44&16.74/6.72\\
\textbf{T5}&18.54/4.08&13.17/2.05&18.98/10.17&28.87/15.48\\
\midrule
\textbf{RA-T5}&18.68/3.64&14.49/4.47&23.98/13.17&31.72/17.97\\
\bottomrule
    \end{tabular}
    \caption{BLEU-1/BLEU-2 for \textsc{st2mo} and \textsc{mo2st} on the validation set.}%We present the results of perplexity just as a reference for future work. 
    %N/A means that there are not public Chinese models currently. 
    %All scores are multiplied by 100. The best performance is highlighted in \textbf{bold}. }%The scores marked with  means the model outperforms the previous one significantly with $p<0.01$~(sign test). Note that we do not conduct the significance test }
    \label{tab:auto_eva_val}
\end{table}

\begin{figure}[!t]
\includegraphics[width=\linewidth]{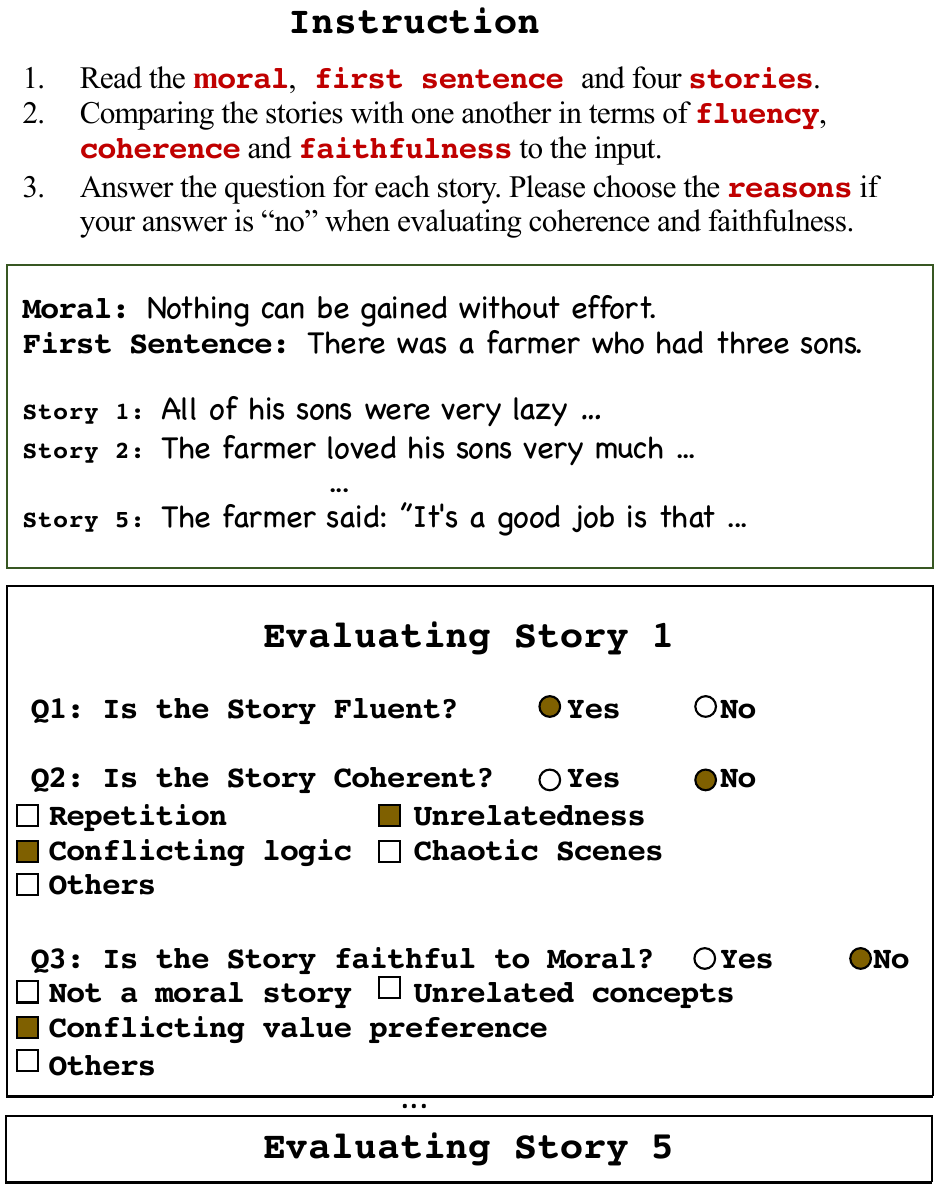}  
\caption{A simplified version of the manual annotation interface for \textsc{mo2st}. The interface for \textsc{st2mo} is similar.}
  \label{fig:ant}
\end{figure}

\iffalse
\begin{table}[!tp]
\small
    \centering
    \begin{tabular}{lccccc}
    \toprule
    {\textbf{Models}}&\textbf{\textsc{rept}}&\textbf{\textsc{relat}}&\textbf{\textsc{cont}}&\textbf{\textsc{chao}}&\textbf{Others}\\
    %&\textbf{Rept}&\textbf{Unrel}&\textbf{Conf}&\textbf{Chao}&\textbf{Not}&\textbf{Concep}&\textbf{}\\
    
    \midrule
    \multicolumn{4}{l}{\textbf{Task: \textsc{st2mo}}}\\    
    \midrule
    \textbf{Fusion}&8\%&35\%&25\%&58\%&6\%\\
    \textbf{T5}&5\%&15\%&23\%&16\%&5\%\\
    \midrule
    \textbf{{Truth}}&0\%&1\%&1\%&1\%&1\%\\
    \midrule
 %   \midrule
%\textbf{Model}&\textbf{M}&\textbf{M}&\textbf{M}&\textbf{M}&\textbf{M}&\textbf{M}&\textbf{M}\\    
\midrule
    \multicolumn{4}{l}{\textbf{Task: \textsc{mo2st}}}\\
    \midrule
    \textbf{Fusion} &11\%&57\%&39\%&44\%&2\%\\
    \textbf{T5}&34\%&30\%&42\%&39\%&2\%\\
    \midrule
    \textbf{{Truth}}&0\%&0\%&0\%&1\%&0\%\\
    
    \bottomrule    
    \end{tabular}
    \caption{Percentage of the texts which are annotated with some error in all annotated 100 texts in terms of coherence. The error types include: repetition~(\textbf{\textsc{rept}}), unrelatedness~(\textbf{\textsc{relat}}), contradictory logic~(\textbf{\textsc{cont}}), chaotic scenes~(\textbf{\textsc{chao}}) and \textbf{Others}.}
    \label{tab:error_cohe}
\end{table}
\fi
\paragraph{Results on Validation Sets}
%Besides the performance on the test set reported in the main paper, 
We show the performance of several baselines and RA-T5 on the validation sets of the understanding tasks and the generation tasks in Table~\ref{tab:select_acc_val} and Table~\ref{tab:auto_eva_val}, respectively.

\subsection{Manual Evaluation Instruction}\label{instruc}
%\paragraph{Annotation Instruction} 
We show the manual annotation interface in Figure~\ref{fig:ant}. To ensure that the annotators guarantee a consistent standard in the annotation process, we asked annotators to rate four examples with the same input at the same HIT~(human intelligence task). In these four examples, one is written by humans and three are generated by models~(i.e., Fusion, T5 and RA-T5). We payed each annotator \$0.2 on average for annotating each example. 
%since Likert-scale based rating has been reported to suffer from low inter-rater agreement~\cite{venkatesh2018evaluating} because raters hardly guarantee to maintain a consistent standard.  %\citet{fang2018sounding} also observed that the distribution of Likert-based ratings heavily skews to the positive end of rating scale. 
%Therefore, the community recently tends to leverage pairwise comparison for manual evaluation~\cite{vinyals2015neural} since selecting a better option is easier for raters than deciding an absolute 

\subsection{Significance of Manual Evaluation Results}\label{signi} 
Table~\ref{tab:man_eva_pvalue} shows the $p$-values~(sign test) when comparing the manual evaluation results~(Table~\ref{tab:man_eva} in the main paper) between each pair of the ground truth, Fusion, T5 and RA-T5.

\begin{table}[ht]
\scriptsize
    \centering
    \begin{tabular}{c|l|c|c|c}
    \toprule
    \textbf{Tasks}&\textbf{Models}&\textbf{Flu}&\textbf{Cohe}&\textbf{Faith}\\
    \midrule
    \midrule
    \multicolumn{5}{c}{\textbf{Dataset: \textsc{storal-zh}}}\\
    \midrule
    \multirow{3}{*}{\rotatebox{90}{\textbf{\textsc{st2mo}}}}&\textbf{T5 vs. Fusion}&8.55e-14&3.82e-11&1.55e-6\\
    %&\textbf{Truth vs. Fusion}&5.05e-29&2.52e-29&1.26e-29 \\
    &\textbf{{RA-T5 vs. T5}}&{0.03}&\textbf{0.75}&\textbf{0.23}\\
    &\textbf{Truth vs. RA-T5}&6.10e-5&2.91e-11&2.35e-14\\
    \midrule
    \multirow{3}{*}{\rotatebox{90}{\textbf{\textsc{mo2st}}}}&\textbf{T5 vs. Fusion}&2.35e-3&2.44e-4&\textbf{0.38}\\
    &\textbf{RA-T5 vs. T5}& \textbf{0.14}&{0.04}&\textbf{0.11}\\
    &\textbf{{Truth vs. RA-T5}}&2.27e-12&1.08e-19&1.1e-24\\
    \midrule
    \midrule
    \multicolumn{5}{c}{\textbf{Dataset: \textsc{storal-en}}}\\
    \midrule
      \multirow{3}{*}{\rotatebox{90}{\textbf{\textsc{st2mo}}}}&\textbf{T5 vs. Fusion}&3.71e-11&2.92e-12&7.92e-9\\
    &\textbf{RA-T5 vs. T5}&\textbf{0.27}&{0.06}&4.18e-3\\
    &\textbf{{Truth vs. RA-T5}}&7.20e-3&4.08e-3&4.92e-5\\
    \midrule
    \multirow{3}{*}{\rotatebox{90}{\textbf{\textsc{mo2st}}}}&\textbf{T5 vs. Fusion}&{0.04}&{0.09}&{0.04}\\
    &\textbf{RA-T5 vs. T5}&\textbf{0.5}&\textbf{0.25}&7.81e-3 \\
    &\textbf{{Truth vs. RA-T5}}&1.46e-11&1.42e-14&1.71e-8\\
    \bottomrule    
    \end{tabular}
    \caption{$p$-values (sign test) when comparing each pair of the ground truth and three models for the manual evaluation results. We highlight the $p$-values larger than 0.1 in \textbf{bold}, which indicates A has an insignificant superiority w.r.t. B for ``A vs. B''.}
    \label{tab:man_eva_pvalue}
\end{table}

\subsection{Evaluating Value Preference Alignment}\label{man_align}
Although we have used \textsc{moPref}  to evaluate whether machines can capture the value preference of a story, the automatically constructed dataset may bias machines to focus on distinguishing general standards of good behaviour without considering story plots. Therefore, in this section, we construct examples manually to test this ability  %investigate whether existing models are able to judge the consistency of value preferences between morals and stories 
beyond the token level. Specifically, we randomly sampled 50 examples from the test sets of  \textsc{storal-zh} and \textsc{storal-en} respectively. For each example, we manually rewrote the moral to convey a synonymous or antonymous value preference. %And we ensured the token-level relatedness between the rewritten morals and the story plot. 
For example, a synonymous moral with ``unity is strength'' in Table~\ref{tab:example_story} can be ``we are powerful as long as we unite with each other'' and an antonymous one can be ``everyone can also be powerful enough.''
Then we expect a model to be able to accept the synonymous moral but reject the antonymous one. We use three typical models, including BERT w/o Story, RA-RoBERTa and RA-T5, to compute the winning rate of pair-wise comparisons between any two of ground-truth, synonymous and antonymous morals. These models are trained on the training set of the \textsc{moPref} task. %We trained RoBERTa as a faithfulness scorer as described in Section~\ref{moral_corr}, and we regard the moral with a higher faithfulness score as the winner in pair-wise comparisons. As for T5, we directly used the model trained on the \textsc{st2mo} task and we regard the moral with a lower perplexity as the winner.

%use the classifier described in section \ref{moral_corr} to judge the faithfulness of the rewritten morals.

\begin{table}[!ht]
\scriptsize
    \centering
    \begin{tabular}{l|c|c|c}
    \toprule
    %\textbf{Models}&\textbf{Truth}&\textbf{Synonymous}&\textbf{Antonymous}\\
    \textbf{Models}&\textbf{True vs. Syn}&\textbf{True vs. Ant} &\textbf{Syn vs. Ant}\\
    \midrule
    \midrule
    \multicolumn{4}{c}{\textbf{Dataset: \textsc{storal-zh}}}\\    
    \midrule
    \textbf{BERT w/o Story}&52\%~(0.89)&46\%~(0.67)&58\%~(0.32)\\
   \textbf{{RA-RoBERTa}}&40\%~(0.21)&36\%~(0.06)&48\%~(0.89)\\
   %\textbf{Ours~({T5})}\\
   \midrule
   \midrule
\multicolumn{4}{c}{\textbf{Dataset: \textsc{storal-en}}}\\
\midrule
    \textbf{BERT w/o Story}&54\%~(0.67)&54\%~(0.67)&48\%~(0.89)\\
   \textbf{RA-RoBERTa}&64\%~(0.06) &34\%~(0.03)&40\%~(0.20)\\
   %\textbf{Ours~({T5})}\\
    \bottomrule
    \end{tabular}
    %\caption{Moral fathifulness scores of the ground-truth, synonymous and antonymous morals.}
    \caption{Winning rates of pair-wise comparisons which require selecting a correct moral from two candidates. Each candidate is a ground-truth~(\textbf{True}), synonymous (\textbf{Syn}), or antonymous (\textbf{Ant}) moral. The number in the parenthesis is the corresponding $p$-value~(sign test).}
    \label{tab:moral_understand}
\end{table}

Table~\ref{tab:moral_understand} shows the evaluation results. We observe that BERT can not distinguish different types of morals without input stories. RA-RoBERTa
%T5 for \textsc{storal-zh} is good at distinguishing synonymous morals from antonymous ones~(winning rate of 72\%), 
fails to accept the synonymous morals on \textsc{storal-en}~(winning rate of only 36\% w.r.t the ground truth, $p<0.1$), and can not distinguish synonymous and antonymous morals on both \textsc{storal-zh} and \textsc{storal-en}~(winning rate near 50\% with $p>0.1$). Additionally, it prefers antonymous morals to the ground truth significantly on both datasets~(winning rate less than 50\% and $p<0.1$ ). The results indicate that existing models still struggle to capture the value preference of moral stories.

\begin{table}[!t]
\small
    \centering
    \begin{tabular}{l|cccc}
    \toprule
    {\textbf{Models}}&\textbf{\textsc{nam}}&\textbf{\textsc{unrel}}&\textbf{\textsc{conf}}&\textbf{Others}\\
    %&\textbf{Rept}&\textbf{Unrel}&\textbf{Conf}&\textbf{Chao}&\textbf{Not}&\textbf{Concep}&\textbf{}\\
    
    \midrule
    \multicolumn{5}{c}{\textbf{Task: \textsc{st2mo}}}\\    
    \midrule
    \textbf{Fusion}&27\%&23\%&7\%&2\%\\
    \textbf{T5}&19\%&9\%&12\%&0\%\\
    \textbf{RA-T5}&15\%&7\%&6\%&0\%\\
    \midrule
    \textbf{{Truth}}&3\%&4\%&2\%&0\%\\
    \midrule
 %   \midrule
%\textbf{Model}&\textbf{M}&\textbf{M}&\textbf{M}&\textbf{M}&\textbf{M}&\textbf{M}&\textbf{M}\\    
\midrule
    \multicolumn{5}{c}{\textbf{Task: \textsc{mo2st}}}\\
    \midrule
    \textbf{Fusion}&25\%&13\%&6\%&1\%\\
    \textbf{T5} &19\%&9\%&10\%&1\%\\
    \textbf{RA-T5}&16\%&10\%&10\%&0\%\\
    \midrule
    \textbf{{Truth}}&2\%&1\%&1\%&0\%\\
    
    \bottomrule    
    \end{tabular}
    \caption{Percentage of the texts annotated with a certain error in all annotated 100 texts in terms of moral faithfulness.}
    \label{tab:error}
\end{table}

\begin{table*}[!t]
\scriptsize
    \centering
    \begin{tabular}{p{443pt}}
    \toprule
    \multicolumn{1}{c}{\textbf{Understanding Task: \textsc{moCpt}}}\\
    \midrule
    \textbf{Input Story:} A stag had fallen sick. He had just strength enough to gather some food and find a quiet clearing in the woods, where he lay down to wait until his strength should return. The animals heard about the stag's illness and came to ask after his health. Of course, they were all hungry, and helped themselves freely to the stag's food; and as you would expect, the stag soon starved to death.\\
    %\textbf{Input Story:} Two goats, frisking gayly on the rocky steeps of a mountain valley, chanced to meet, one on each side of a deep chasm through which poured a mighty mountain torrent. The trunk of a fallen tree formed the only means of crossing the chasm, and on this not even two squirrels could have passed each other in safety. The narrow path would have made the \textit{\textbf{{bravest}}} tremble. Not so our goats. Their pride would not permit either to stand aside for the other. One set her foot on the log. The other did likewise. In the middle they met horn to horn. Neither would give way, and so they both fell, to be swept away by the roaring torrent below.\\
    \midrule
    \textbf{Candidate Moral 1:} Good will is worth nothing unless it is accompanied by good acts. \newline
    \textbf{Candidate Moral 2:}  Every man in need is your neighbor. \newline
    \textbf{Candidate Moral 3: }Your everyday good deeds never go in vain as they will return to you when you least expect them. \newline
    \textbf{Candidate Moral 4:} Don't trust strangers.  \newline
    \textbf{Candidate Moral 5:} Everyone person is significant and deserve your attention and respect.\\
    \midrule
    \textbf{True Answer:} Moral 1\newline\textbf{Model Prediction:} Moral 5\\
    \midrule
    \midrule
    \multicolumn{1}{c}{\textbf{Understanding Task: \textsc{moPref}}}\\
    \textbf{Input Story:} Once upon a time there lived a cat that loved to read. At night, when everybody was asleep, she would put on the spectacles and read the big book for cats. One day, she read in the book: if you want a mouse for dinner, repeat the following rhyme: in this house there is a mouse, where is the mouse, where is the mouse? The cat looked up from the book and found that there was a mouse on the top of the table. The cat repeated the rhyme and soon found the same mouse on the bed. Then she jumped upon the bed to catch the mouse and the mouse was gone! The mouse was very clever. Suddenly he squeaked, "Oh, dear cat, run, run fast! there is dog after you!" The cat left the mouse and was ready to jump out of the window. The mouse sat near his hole and said, "Ha-ha-ha! dear cat that was the trick I learnt from the bio book for mice!" And the mouse ran into his hole!\\
    \midrule
    \textbf{Candidate Moral 1:} An intelligent person should not think that others are illiterate.\\
    \textbf{Candidate Moral 2:} An intelligent person should not forget that others are illiterate.\\
    \midrule
    \textbf{True Answer:} Moral 1\\
    \textbf{Model Prediction:} Moral 2\\
    \midrule
    \midrule
    \multicolumn{1}{c}{\textbf{Generation Task: \textsc{st2mo}}}\\
    \midrule
    \textbf{Input Story:} In the forest, there was a deer and an owl. The deer loved to scare this fearful owl all the time. One day, the deer scared the owl so hard, the owl past out. The deer ran away, and the owl woke up and came to her senses. She had had enough of being scared. She went to the deer and said, ``I'm not scared of you.'' She said this with such a force that the deer backed down $\cdots$\\
    \midrule
    \textbf{Generated Case 1:} {\underline{The deer} loves to scare \underline{owls}.} \textsc{(nam)}\\
    \midrule
    \textbf{Generated Case 2:} {We should not be \underline{greedy} and learn to others the importance of our \underline{desires}.} \textsc{(unrel)}\\
    \midrule
    \textbf{Generated Case 3:} {It's \underline{good} to be scared.} \textsc{(conf)}\\
    %\midrule
    %\textbf{Case4:} We should not let fear control us.\\
    \midrule
    \textbf{True Moral:} Don't let anyone scare you all the time.  \\
    \midrule
    \midrule
    \multicolumn{1}{c}{\textbf{Generation Task: \textsc{mo2st}} }\\
    \midrule
    \textbf{Input Moral:} Empty solutions are of no worth.  \\
    \midrule
    \textbf{Generated Case 4:} \textbf{There was a grocery shop in a town. } There was \textbf{a big fat cat} in there. It was very big. It was a giant cat. It was big, and it was very big and powerful. It was big and powerful. It was very powerful. The mouse could \textbf{move freely} and had a nice time hunting. The cat was very big and powerful. The mouse wanted to live in a very large land $\cdots$ \textsc{(nam)}\\%. The mouse was very big and strong.\\
    \midrule
    \textbf{Generated Case 5:} \textbf{There was a grocery shop in a town. } %Just the mouse and cat lived together. They were doing very well. It was very busy. 
    Just the mouse and cat lived together. $\cdots$ The mouse was very \underline{active} in the shop and in the street. The mouse was very happy. But the mouse was too \underline{busy} to \textbf{move freely}. He was the only mouse in town. He was very \underline{busy} with work. $\cdots$ %He could not move freely. When he went to the cat, he fell into the street. He was so happy. He kept going. One day, there was a big fat cat in the street. The cat was so quiet. The mouse was so quiet and he was very quiet. The cat moved slowly. The cat moved softly. As soon as the mouse passed his way, he was happy to see the mouse. The mouse was so happy that the cat was very happy. The cat was very sad. 
    As soon as the mouse had left, he moved slowly. $\cdots$ %The mouse was so happy. The cat was very happy. The mice had to give up his life. Seeing this, the cat moved softly and again he was able to move freely. The cat moves softly, so the mouse was very happy. The mouse was so happy that the mouse was happy. When the cat was so happy, the mouse was very happy and happy. The cat walked slowly beside the cat, but the mouse would not be able to move freely. The cat was very happy to have a nice time hunting with the mouse and moved freely and was so happy to have a little bit of good food for the mice. 
    So the mouse was so happy and \underline{satisfied} with everything that it was so \underline{happy} for him. $\cdots$~\textsc{(unrel)}\\%The mouse and cat sat happily. The mouse and cat moved freely. The mice were so happy for him. The mouse and cat moved freely. The cat moved slowly, and the mouse and the cat walked happily at the same time hunting the mouse. The mouse sat still there for the cat. The cat moved to the next place. They had already crossed the road. The mice walked in the grass. The mice had a nice time hunting the mice. The mice wanted to live there and ate the cat and the mice ate the cat. The cat seemed to be able to play, and the cat ate the mouse. The cats wanted to play with the mouse. The mice wanted to play with the cat. The cat moved slowly stood beside the mouse. The mouse had a nice time huntin\\
    \midrule
    \textbf{Generated Case 6:} \textbf{There was a grocery shop in a town. } There were two stores. $\cdots$ %Each was sold separately. The first was a big fat cat. The second was a mouse. The third was a mouse. The mouse was very happy. Then a mouse was born. The first was a mouse and he loved him. The first was a mouse. The second was a cat. The third was a cat. The second was a cat. The first was a mouse. The third was a cat. The third was a mouse. The third was a big fat cat. The third was a mouse. They were very happy. The third was a mouse who sat down and sat down and sat with the fourth mouse. The third was a cat to move softly and slowly. The third was a cat. The third was a mouse. It was a mouse. The fourth was a big fat cat. The third were a mouse. The fourth was a cat. Then the fifth was the mouse. The third was a nice time hunting the third. The fifth was a big fat cat. The last was a mouse. The third was a big fat mouse. To be a nice time hunting, the fourth was a big fat cat. The fifth was the mouse. The third was a big fat cat. The fourth was a mouse. Finally, the fourth was a mouse. The fifth was a cat. The third was a mouse. The fourth was a big fat cat. The fifth was a cat. 
    \underline{The third was an empty solution}. Each time the mouse went on a few days. Then, the third was a little while later the next day the mice were gone. They were gone. \underline{They were happy at last} $\cdots$ \textsc{(conf)}
    %, and they had lost their happiness. They saw that the next day the next time the mice were happy with their new solution was to move freely. The second was a very big fat cat. The fourth was a big fat cat. The last part was a big fat cat. The third was a mouse. Then a cat went to the mouse. Then, as the second was a big fat cat moved softly. Finally, the third was a cat. The third was the last mouse. The third was the first and the third was a big fat cat. The last time the mice wanted to g
\\
    \midrule
    \textbf{True Story:} \textbf{There was a grocery shop in a town. }Plenty of \textbf{mice lived} in that grocery shop. Food was in plenty for them. They ate everything and spoiled all the bags. They also wasted the bread, biscuits and fruits of the shop. The grocer got really worried. So, he thought ``I should buy a cat and let it stay at the grocery. only then I can save my things.'' He bought a nice, \textbf{big fat cat} and let him stay there. The cat had a\textbf{ nice time hunting} the mice and killing them. The mice could not \textbf{move freely} now. They were afraid that anytime the cat would eat them up. The \textbf{mice wanted} to do something. They held a meeting and all of them tweeted ``We must get rid of the cat. can someone give a suggestion''? All the mice sat and brooded. A smart looking \textbf{mouse stood} up and said, ``The\textbf{ cat moves softly.} that is the problem. if we can tie a bell around her neck, then things will be fine. we can know the movements of the cat''. ``Yes, that is answer,'' Stated all the mice. An old \textbf{mouse slowly stood} up and asked, ``Who would tie the bell?'' After some moments there was no one there to answer this question. \\
    \bottomrule    
    \end{tabular}
    \caption{Typical error cases predicted by RA-T5~(for the understanding tasks) or sampled from RA-T5~(for the generation tasks). For the generation tasks, the error types in terms of moral faithfulness include ``not a moral text''~(\textsc{nam}), ``unrelated concepts''~(\textsc{unrel}) and ``conflicting value preference''~(\textsc{conf}). The \underline{underlined} words are improper concepts/events which leads to corresponding errors.  \textbf{Bold} words for \textsc{mo2st} are the given first sentence and the outline of multiple phrases. }
    \label{tab:case}
\end{table*}

\begin{table*}[!t]
\scriptsize
    \centering
    \begin{tabular}{p{443pt}}
\toprule
\multicolumn{1}{c}{\textbf{Generation Task: \textsc{st2mo}}}\\
\midrule
\textbf{Input Story:} Once upon a time there was a spring who lived happily and safely inside a pen. Although he heard many noises coming from outside, he lived believing that outside his world inside the pen, there was nothing good. Even just to think about leaving his pen made him so scared that he was quite content to spend his life compacting and stretching himself again and again inside that tiny space. However, one day, the ink ran out, and when the pen's owner was busy changing it, there was an accident. The spring was flung through the air and landed in the toilet drain, well out of sight. Terrified, and cursing his bad luck, the spring was flushed through pipe after pipe, each time thinking it might be his end. During the journey, he did not dare open his eyes out of pure fear. Nor did he every stop crying. Swept away by the water, he travelled on and on and on, until he ended up in a river. When the river current lost its force, and the spring could see that things had calmed down a bit, he stopped crying and listened all around him. Hearing birdsong and wind in the trees, he felt encouraged to finally open his eyes. What the spring saw was the pure, crystal waters of the river, the rich green rocks of the riverbed, and all kinds of fish of many colours, whose skin seemed to dance under the sunlight. Now he understood that the world was much greater than the space inside the pen, and that there had always been many things outside, waiting to be enjoyed. After spending a while playing with the fish, he went over to the riverbank, and then moved on to a field of flowers. There he heard weeping. He followed the sound, which took him to a lovely flower that had been flattened by a rabbit, and could no longer stand up straight. The spring realised that he could help the flower, so he offered to be his support. The flower accepted, and slipped through the middle of the spring. There they lived happily together. And they would always laugh when remembering how the spring used to think that all there was to life was being a sad and fearful spring.\\
\midrule
%\textbf{Fusion:} Never follow to make yourself and cares in life in one or right attitude and think.~(\textsc{unrel})\\
\textbf{GPT2:} Always \underline{think} of what you can do to make things better.\\
\midrule
\textbf{T5:} Fear and \underline{frustration} are one thing, but they can be turned into happiness.\\
\midrule
\textbf{RA-T5:} Life should not be afraid of the unknown. There should be lots of positive things out there.\\\\
\textbf{Top-5 Retrieved Morals by RA-T5:} \\
(1) Everything in life has its unpleasant or uncomfortable side. These aspects cannot be removed from life. \\(2) It's always better to have friends and be happy - even though it may bring discomfort - than isolate yourself from people in order to avoid problems. \\{{(3) \hlred{Most fear is without foundation, and plays on what is unknown or different.}}} \\(4) Friends are for cheering you up in your worst moments. \\(5) Someone who harm's no one is still very far from being good.%(8) Often rumours and idle gossip are based on nothing but speculation, and spread without being questioned. (9) Things change, and we have to adapt to this and make the most of the new opportunities this brings; even when that means renouncing what we already have. (10) It's important to keep things clean.
\\
\midrule
\textbf{True Moral:} Changes are not a bad thing, it is up to ourselves to open the eyes and make something positive of them.\\
\midrule
\midrule
\multicolumn{1}{c}{\textbf{Generation Task: \textsc{st2mo}}}\\
\midrule
\textbf{Input Moral:} Never give up. \\
\midrule
%\textbf{Fusion:} \textbf{One day a dragon and a dino met in the late jurassic period.} In the class the bear began to fight a big box. So the king decided to play in the desert. The bear went to the donkey to pick him even in the house. Then he took another meal and decided to go to the window when the cheetah got enough for his life and he wanted to make a job to a parrot. Then he gave the boy to the room, and he felt out of the wn's house. Then the king felt imeasy and he could not do to tell his play with the does. He was so surprised to see that his friend was a good hand with a bird and put his tail to the king. Then he gave his wife where he was still all the animal in the window, the donkey looked inside. Then he asked the boy to see the old man.\\
\textbf{GPT2:} \textbf{One day a dragon and a dino met in the late jurassic period.} They were friends, and on the way they met a \textbf{huge tree}. Dragon was huge and his \textbf{big claws} were full and he was going to kill him. \underline{Dino was going to die. Dino was going to die.} When the \textbf{dragon made half} of the tree, he was about to die. Dragon was going to die, and all the \underline{beasts and beasts} were dead. Dino had no idea what to do. He was the \textbf{strongest dino} in the whole tree. Dragon was a \textbf{total surprise}. Dino was going to die. \underline{He was about to die. He was going to die.} The dragon was astonished. Dino was not able to die. Dino was so angry, he started to die. Dino was so angry, and the dragon threw his \textbf{big claws} at the tree and ran to the forest. This was the best match. Then the dino was nearly killed.\\
\midrule
\textbf{T5:} \textbf{One day a dragon and a dino met in the late jurassic period.} The dragon was standing next to a \textbf{huge tree}. It was full of \underline{beasts, lions and zebras} that were strong enough to carry out their duties. The dragon was so strong that he could move himself without losing his foot. The dino was so strong that he could carry anything, even if that tree had much more branches. But the dino was very weak, so he could not carry anything. The dragon came to the tree and said, ``I will do everything I can to get my foot into the tree.'' Then the \textbf{dino started running} through the tree. The \textbf{dragon made half} way through the tree and put on his \textbf{big claws}. Then he took off his \textbf{big claws} \underline{in his claws}. Just then a huge \textbf{tree started} to go through a storm. Then the monster took out his \textbf{fiery breath}. The dino could not contain his \textbf{fiery breath} and flew away.\\
\midrule
\textbf{RA-T5:} \textbf{One day a dragon and a dino met in the late jurassic period.} The dragon was very proud of the dino because of his \textbf{big claws} and \textbf{fiery breath}. The dino was going to jump on a \textbf{huge tree} and climb it up. But the dragon didn't know how to jump. The \textbf{dragon made half} way up the tree and \hlred{started} to roll down. The dino was completely surprised. The dragon said, ``I can't jump,'' The dragon started to jump. Then the \textbf{tree \hlred{started}} to roll. The dino \hlred{started} to roll over the \hlred{top of the tree} and jumped up. The \textbf{tree \hlred{started}} to thrash off. \hlred{Suddenly}, the dino saw that the dragon had a \textbf{fiery breath}. The dragon said, ``That's my strength.'' The dino was all alone. Then he stood, looking at the dragon. Then the dragon stopped, shouted at him, ``Hello dino! what a \textbf{total surprise}.'' The \textbf{dino \hlred{started} running} and jumping towards the tree, and the dragon got close to him. Then he hit the tree. Now he found a boulder on the tree \hlred{started} to crash and he jumped on the tree. Then the \textbf{strongest dino} ever was {surprised}.\\\\
\textbf{Top-5 Retrieved Story Outines:} \\
(1) \{baldwin flew, baldwin scratched rattler, team beat baldwin, baldwin dodged rattler, football game, goal post, baldwin threw rattler, baldwin \hlred{started}\}\\
(2) \{eagle resting, \hlred{tree top}, tortoise rested, eagle answered, deep sleep, tortoise sleeping, tortoise smiled, hunter suddenly]\}\\
(3) \{loud thump, man happened, cry intruded, ugly wolf, wolf named pete walked, long neck, man walked, thin air\} \\
(4) \{cat \hlred{suddenly} fell, bird flew, \hlred{started} climbing, cat thanked\}\\
(5) \{lion won, race \hlred{started}, croc won\}\\
\midrule
\textbf{True Story:} \textbf{One day a dragon and a dino met in the late jurassic period.} The dragon said, ``I'm stronger than you!'' The dino said, ``I'm the \textbf{strongest dino} ever!'' The next day the dino and the dragon met in the forest. The \textbf{dragon made half} of a tree fall down with its \textbf{big claws} and \textbf{fiery breath}. The dragon said, ``You can't beat that!'' The \textbf{dino started running} toward a \textbf{huge tree}. The dino rammed the huge tree with its head. Nothing happened. The dragon laughed. Then the \textbf{tree started} to fall. The dragon just stared in \textbf{total surprise}.\\
\bottomrule
    \end{tabular}
    \caption{Cases generated by different models for the generation tasks. %The error types in terms of moral faithfulness include ``not a moral text''~(\textsc{nam}), ``unrelated concepts''~(\textsc{unrel}) and ``conflicting value preference''~(\textsc{conf}). 
    The \underline{underlined} words are improper concepts/events which leads to incoherence or unfaithfulness. \textbf{Bold} words for \textsc{mo2st} are the given first sentence and the outline of multiple phrases. The \hlred{red} moral for \textsc{st2mo} is related to the generated moral of RA-T5 in semantics. Note that we only take concepts in these retrieved morals as input for RA-T5. And \hlred{red} words in the retrieved outlines for \textsc{mo2st} indicate that they also show up in the generated story of RA-T5.}
    \label{tab:case_good}
\end{table*}

\section{Error Analysis and Case Study}\label{case_error_study}
In this section, we conducted a case study and investigated the errors of existing models on the proposed tasks to provide insight into future work. We show several typical error cases in Table~\ref{tab:case}. 
\subsection{Understanding Tasks} The example in Table~\ref{tab:case} for \textsc{moCpt} shows that the model may not grasp abstract concepts such as ``good will'' and ``good acts'' and align them to the story plots. It makes predictions possibly based on only token-level features such as relations between ``ask after'' and ``attention''. On the other hand, the example for \textsc{moPref} indicates that the model can not capture the value preference of the story in terms of ``whether it is intelligent to regard others are illiterate''. The results demonstrate the necessity of introducing concept knowledge and modeling high-level semantic information.

\subsection{Generation Tasks} Table~\ref{tab:case_good} shows cases generated by several baselines and our model for the generation tasks. We can see that retrieval can provide effective guidance for both moral and story generation. Baseline models including GPT2 and T5 tend to generate unrelated concepts or non-moral texts.

However, as shown by the manual evaluation results, there is still a big gap between RA-T5 and humans. To provide quantitative error analysis, in the process of manual evaluation on \textsc{storal-en}, we required annotators to annotate the error type of a text when it exhibit an unfaithful moral. We summarize three main error types as follows: \textbf{(1) Not a moral text~(\textsc{nam})}: not stating or implying what is right or what is wrong; \textbf{(2) \textbf{Unrelated concepts~(\textsc{unrel})}}: containing unrelated concepts with the input; and \textbf{(3) Conflicting value preference~(\textsc{conf})}: conveying a value preference conflicting with the input despite related concepts. In addition, we also provide annotators with another option \textbf{Others}. %Although the error types may be not independent, 
The annotators are allowed to annotate a text with multiple errors. %based on subjective feelings. W
When at least two of three annotators annotate the text with some error, we decide it has the error. %when at least two of three annotators annotate the text with the error. 
We show the distribution of the error types in Table~\ref{tab:error}, suggesting that existing models still struggle to generate meaningful morals and stories, and align the concepts and value preferences between them. 

Furthermore, as exemplified in Table~\ref{tab:case},
%we present typical cases generated by the evaluated models for each error type in the appendix. %and the error analysis in terms of incoherence.
%\subsection{Philosophy Understanding and Generation}
%Story might have philosophies, whether it is a concept or exhortation. Philosophies are different to summaries: summaries are condensation of contents, but philosophies are derivations from contents, which requires higher level comprehension in stories. To better evaluate models' properties in understanding and generation, we design two tasks: philosophy extraction and generation from given philosophy. 
%\paragraph{Case Study}
%We present typical cases generated by ours~(T5) for each error type in terms of moral correctness in Table~\ref{tab:case}. 
when generating morals, we can see from Case 1 that the models still often state events involved with specific characters~(e.g., \textit{``owls''}) but do not tell what is right and what is wrong. And Case 2 shows that they struggle to conclude related concepts from the story~(e.g., \textit{``greedy''} is not embodied in the story at all). Furthermore, in Case 3, the models conclude a conflicting value preference with the story despite correct concepts~(e.g., the story shows that \textit{``it is bad to be scared''} but not \textit{``good''}). On the other hand, models also are shown to suffer from similar issues when generating stories. In Case 4, the model only describes some scenes~(e.g., \textit{``it was very big''} and \textit{``it was very powerful''}) but does not aims to convince readers of anything. And Case 5 seems to tell a story centered on some concepts such as \textit{``active''} and \textit{``busy''}, but the concepts do not relate to the input. Case 6 implies \textit{``empty solutions may be useful,''} which is conflicting with the input. These cases indicate the necessity of modeling the relations between events and abstract concepts for understanding and generating  moral stories.

\end{document}